\documentclass[twoside]{article}

\usepackage[accepted]{aistats2023}
\usepackage[round]{natbib}

\usepackage[dvipsnames]{xcolor}

\usepackage{graphicx}

\usepackage{booktabs}
\usepackage{caption}
\usepackage{multirow}

\usepackage{amsmath}
\usepackage{amssymb}
\usepackage{amsthm}
\usepackage{mathtools}

\usepackage{hyperref}
\hypersetup{
    colorlinks=true,
    linkcolor=Blue,
    citecolor=Blue,
    urlcolor=Blue
    }
\usepackage[capitalise,noabbrev]{cleveref}

\def\eqref#1{equation~\ref{#1}}

\def\1{\bm{1}}

\def\rvs{{\mathbf{s}}}

\def\rvx{{\mathbf{x}}}

\def\rvy{{\mathbf{y}}}

\DeclareMathAlphabet{\mathsfit}{\encodingdefault}{\sfdefault}{m}{sl}
\SetMathAlphabet{\mathsfit}{bold}{\encodingdefault}{\sfdefault}{bx}{n}

\def\gN{{\mathcal{N}}}

\newcommand{\R}{\mathbb{R}}

\newcommand{\KL}{D_{\mathrm{KL}}}

\DeclareMathOperator*{\argmax}{arg\,max}
\DeclareMathOperator*{\argmin}{arg\,min}

\DeclareMathOperator*{\E}{\mathbb{E}}

\newcommand{\g}{\,|\,}

\newtheorem{lemma}{Lemma}

\begin{document}

\twocolumn[

\aistatstitle{Don't be fooled: label leakage in explanation methods and the importance of their quantitative evaluation}

\aistatsauthor{ Neil Jethani* \And Adriel Saporta* \And  Rajesh Ranganath }

\aistatsaddress{ Grossman School of Medicine, Courant Institute\\New York University \And  Courant Institute\\New York University \And Courant Institute, Center for Data Science\\New York University } ]

\begin{abstract}
Feature attribution methods identify which features of an input most influence a model's output.
Most widely-used feature attribution methods (such as SHAP, LIME, and Grad-CAM) are ``class-dependent'' methods in that they generate a feature attribution vector as a function of class.
In this work, we demonstrate that class-dependent methods can ``leak'' information about the selected class, making that class appear more likely than it is.
Thus, an end user runs the risk of drawing false conclusions when interpreting an explanation generated by a class-dependent method.
In contrast, we introduce ``distribution-aware'' methods, which favor explanations that keep the label's distribution close to its distribution given all features of the input.
We introduce SHAP-KL and FastSHAP-KL, two baseline distribution-aware methods that compute Shapley values.
Finally, we perform a comprehensive evaluation of seven class-dependent and three distribution-aware methods on three clinical datasets of different high-dimensional data types: images, biosignals, and text.
\end{abstract}

\section{INTRODUCTION}
Post-hoc feature attribution methods, which identify the features of an input that most influence predictions, are critical in high-stakes contexts such as healthcare.
Feature attribution methods are used not only to interpret individual predictions, but also to better understand a model's global behavior for model development, knowledge discovery, and quality improvement and assurance.
For example, such methods have been used to detect spurious signals in hip fracture radiographs \citep{badgeley2019deep}, to discover novel gene expression signatures \citep{janizek2021uncovering}, and to identify brain regions that help distinguish between possible sources of dementia \citep{iizuka2019deep}.

Most widely-used feature attribution methods (such as SHAP \citep{lundberg2017unified}, LIME \citep{ribeiro2016should}, and Grad-CAM \citep{selvaraju2016grad}) are ``class-dependent'' methods, which we define to be any approach that generates a feature attribution vector as a function of class. However, we theoretically and empirically show  that class-dependent methods can ``leak'' information about the selected class, making that class appear more likely than it is. Thus, an end user runs the risk of drawing false conclusions interpreting an explanation generated by a class-dependent method.

As an alternative, we define a ``distribution-aware'' method (such as REAL-X \citep{jethani2021explain}) to be a class-independent method that creates explanations based on the change in the label's distribution when the features are perturbed, with a preference for explanations with a small change in distribution. Preferring explanations that keep the label's distribution close to its distribution when given full knowledge of the features ameliorates the miscalibration that can occur when using class-dependent methods.
Further, we consider the evaluation strategy that progressively includes only the top $n$\% of features for each data point and then plots the resulting 
model performances on an inclusion curve \citep{arras2017explaining,petsiuk2018rise,jethani2022fastshap}. For this evaluation strategy, we demonstrate that the optimal feature attribution method is distribution-aware. Finally, we propose a strategy for evaluating a feature attribution method given a fixed model.

In summary, our six primary contributions are the following.
(1) We introduce and define the difference between class-dependent and distribution-aware feature attribution methods.
(2) We demonstrate that explanations generated by class-dependent methods using the true label can leak information about the true label, leading to inflated performance metrics for class-dependent methods, whereas this cannot occur with class-independent methods.
(3) We show that explanations generated by class-dependent methods using the predicted label can leak information about the predicted class, making the predicted class appear more likely than it is.
(4) We establish that the optimal feature attribution vector, as measured by the above evaluation metric, is distribution-aware.
(5) We present two distribution-aware feature attribution methods, SHAP-KL and FastSHAP-KL, that estimate Shapley values, are easy to optimize, and can serve as baselines to facilitate the development of additional distribution-aware methods.
(6) We perform a comprehensive evaluation of seven class-dependent and three distribution-aware feature attribution methods on three clinical datasets of different high-dimensional data types: images, biosignals, and text.

\section{RELATED WORK}
\label{sec:related_work}
Feature attribution methods generally fall into one of two categories, which we review below: removal-based methods and gradient-based methods. See \cref{appendix:method_groups} for relevant feature attribution methods grouped by type.

\paragraph{Removal-based feature attribution methods.}
Removal-based methods remove subsets of the input features to determine their influence \citep{covert2021explaining}.
Many removal-based methods, such as LIME \citep{ribeiro2016should} and SHAP \citep{lundberg2017unified}, perform the removal operation for each sample of data, which can be computationally intensive.
Amortized approaches—such as L2X \citep{chen2018learning}, INVASE \citep{yoon2018invase}, REAL-X \citep{jethani2021explain}, and FastSHAP \citep{jethani2022fastshap}—represent a new form of removal-based explainability that performs the removal operation across multiple samples of data at a time in order to learn models that produce explanations for a sample of data with a single forward pass \citep{fong2017interpretable,schwab2019cxplain}.

Recent work has shown that when using removal-based methods, replacing the removed features with reference values shifts the input out-of-distribution or off-manifold, which can affect explanation quality and make it easier for adversarial attacks \citep{frye2020shapley,slack2020fooling,jethani2022fastshap}.
In addition, some amortized explanation methods, such as L2X and INVASE, can produce explanations that encode the label directly in the shape of the explanation rather than with the feature values the explanation highlights \citep{jethani2021explain}.

\paragraph{Gradient-based feature attribution methods.}
Gradient-based methods determine feature importance using gradients with respect to either the input or intermediate representations of the input \citep{ancona2019gradient}. SmoothGrad \citep{smilkov2017smoothgrad}, for example, measures how sensitive the model output is to small changes in a given feature.
Integrated Gradients (IntGrad) \citep{sundararajan2017axiomatic}, on the other hand, computes the average gradient to measure the salience of input features relative to a user-selected reference input.
Another popular method, Grad-CAM \citep{selvaraju2016grad}, computes the gradient of a class with respect to an intermediate layer of a convolutional neural network (CNN).

Gradient-based methods have been shown to be sensitive to small changes or distributional shifts in the input. For example, adding a constant shift to the input can dramatically change the explanations produced by gradient-based methods \citep{kindermans2019reliability,ghorbani2019interpretation}.
Gradient-based methods can also produce explanations that appear invariant to model parameter and training label randomizations \citep{adebayo2018sanity}.

\section{EVALUATION OF FEATURE ATTRIBUTION METHODS}
\label{sec:evaluation}
A feature attribution method generally produces a single attribution vector that assigns a score to each input feature, where a higher score implies a larger relationship to an output.
For a given data point, a single attribution vector could produce many possible \emph{explanations}, where an explanation is some subset of the features based on the scores assigned by the feature attribution method. For example, one could choose the features with the top one, five, or ten percent of scores.

In order to evaluate a feature attribution method, one could compare its explanations to human benchmark explanations. However, human explanations can be time-consuming and expensive to obtain, 
or may not be available at all.
For example, while a neural network is able to predict diabetes from an electrocardiogram (ECG), it is not yet clear to practitioners what information in the signal is predictive of the disease \citep{jethani2022diabetes}.

Multiple strategies have been proposed for evaluating feature attribution vectors without human benchmark explanations.
One standard evaluation strategy is to progressively include only the top $n$\% of features for each data point and measure the resulting effect on model performance \citep{bach2015pixel,samek2016evaluating,hooker2019benchmark,sturmfels2020visualizing}.
The expectation is that the better a feature attribution method is, the more model performance will improve upon inclusion of only the top-scoring features. 
Model performance using each top $n$\% subset of features is then plotted as an inclusion curve \citep{arras2017explaining,petsiuk2018rise,jethani2022fastshap}.
We follow this evaluation strategy, as described below.

\begin{figure*}[t]
    \centering
    \includegraphics[width=.9\linewidth]{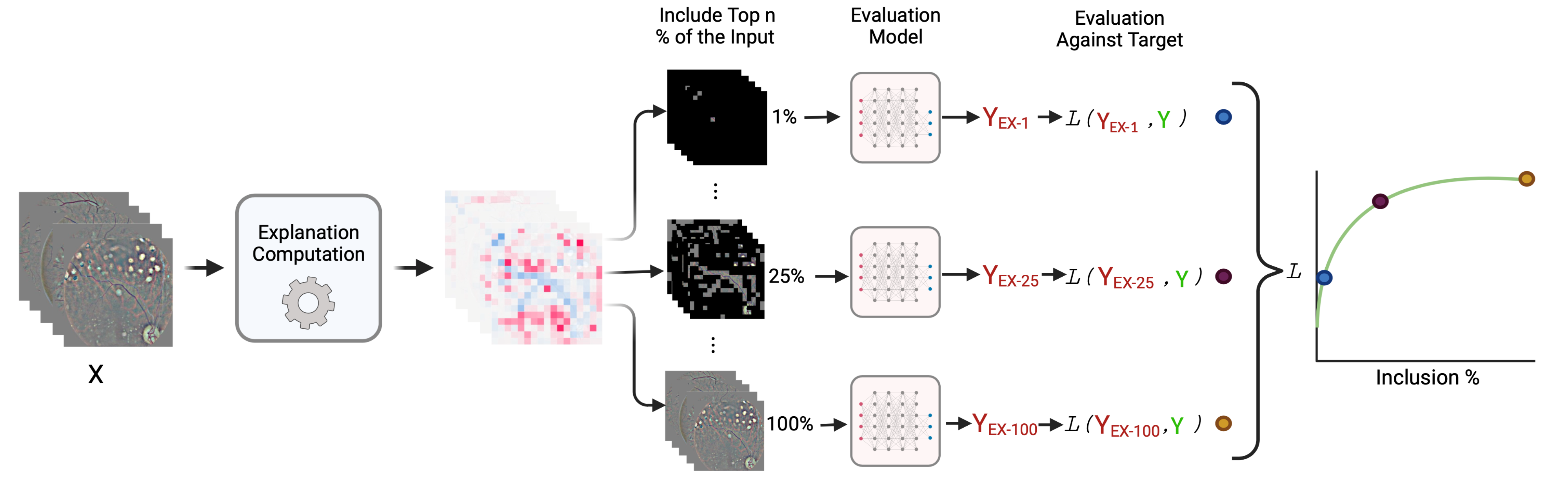}
    \caption{Illustration of the evaluation framework. An inclusion curve is constructed by progressively increasing $n$ from 0 to 100, selecting the top $n\%$ of features for each data point in a held-out test set using the corresponding feature attribution vector, and then measuring performance of the surrogate evaluation model across the entire test set using the log-likelihood.}
    \label{fig:evaluation_fig}
\end{figure*}
\label{sec:eval_definition}
\paragraph{Defining the evaluation.}
Let $\rvx \in \mathcal{X}$ be a random vector consisting of $d$ features, or $\rvx = (\rvx_1, \ldots, \rvx_d)$. Let $\rvy \in \mathcal{Y} = \{1, \ldots, K\}$ be the target variable for a multiclass classification problem.
We use $\rvs \in \{0, 1\}^d$ to denote subsets of the indices $\{1, \ldots, d\}$.
The symbols $\rvx, \rvy, \rvs$ are random variables, and the symbols $x, y, s$ are possible values for those random variables.

\citet{hooker2019benchmark} noted that when relevant features are removed, the new altered input comes from a distribution that is different from that of the original unaltered input, thereby making it difficult to know whether any degradation in model performance is caused by the removal of relevant features or by the shift in distribution. 
The authors solve this problem by training new \emph{surrogate} models on the altered inputs, but it has been shown that this retraining procedure not only is computationally expensive because it requires re-training for each type of explanation, but also allows the surrogate models to incorrectly assign high scores to feature attribution methods that encode the label in the locations of the removed features as opposed to their actual values \citep{jethani2021explain,leemann2022eval}.
In order to prevent the surrogate model from co-adapting to the explanations, recent work has proposed a computationally efficient strategy that trains a single surrogate model with randomly masked inputs \citep{jethani2021explain,jethani2022fastshap,covert2021explaining}. We follow this strategy as described below.

Let $F(\rvx,\rvy)$ be the data distribution from which data is drawn, and let $p(\rvs)$ be the distribution over $\rvs$ where all subsets occur with non-zero probability.
The surrogate evaluation model $p_{\text{surr}}$ is trained to predict the label $\rvy$ given a vector of masked features. Masking is accomplished with a function $m(x, s)$, where the masking function $m$ replaces features $x_i$ where $s_i = 0$ with a \texttt{[mask]} value that is not in the support of $\rvx_i$. The \textit{Surrogate Objective} is
\begin{align}
\label{eqn:surrogate_objective}
    \mathcal{L}&(\beta) = \nonumber \\
    &\E_{F(\rvx)}\E_{p(\rvs)} 
      \Big[ \KL \big( F\left(\rvy \g x\right) \mid\mid p_{\text{surr}}(\rvy \g m(x, s); \beta) \big) \Big],
\end{align}
where $\KL$ is the Kullback–Leibler (KL) divergence. The surrogate model at optimality matches the conditional probability distribution of the target variable given some subset of features. More formally, if $x_s$ is the set $\{x_i : s_i = 1\}$, then $p_{\text{surr}}(y \g m(x, s); \beta) = F(y \g x_s)$ \citep{jethani2021explain, covert2021explaining}.

After training, the surrogate evaluation model can evaluate any feature attribution method. 
Let $e(x,y) \in \R^d$ be a feature attribution vector generated by a feature attribution method for each paired sample of data $x, y$ where $e_i(x,y) \in \R$ is a score for the feature $x_i$.
Let $\texttt{top}_n(e) = \argmax_{s} s^T e \text{, such that } s \in \{0, 1\}^d,\ \|s\|=\lceil \frac{nd}{100} \rceil,\ \text{and } n \in [0,100]$, define an operation that returns an explanation that denotes the top $n$\% of features with the highest attributions $e_i \in \R$.
An inclusion curve is constructed by progressively increasing $n$ from 0 to 100, selecting the top $n$\% of features for each data point in a held-out test set using the corresponding feature attribution vector $e(x,y)$, and then measuring performance of the surrogate evaluation model $p_{\text{surr}}\Big(\rvy \g m\big(x, \texttt{top}_n(e(x,y)) \big);\beta\Big)$ across the entire held-out test set using the log-likelihood.
The area under the inclusion curve (iAUC) is
\begin{multline}
\label{eqn:iauc}
    \text{iAUC} = \E_{n \sim \text{Unif}(0,100)} \E_{ F(\rvx, \rvy)}\\
    \Bigg[\log p_{\text{surr}}\Big(y \g 
    m\big(x, \texttt{top}_n\left(e(x,y)\right); \beta \big)
    \Big)\Bigg].
\end{multline}
A higher iAUC indicates a higher likelihood of the labels averaged across different feature subset sizes.
See \cref{fig:evaluation_fig} for a diagram of the evaluation procedure.

\section{CLASS-DEPENDENT VS. DISTRIBUTION-AWARE METHODS}
\label{sec:class_dist}
In this section, we draw a distinction between class-dependent and distribution-aware feature attribution methods. This new categorization of feature attribution methods exposes an important limitation of class-dependent methods, which are more commonly used than distribution-aware methods. First, we define class-dependent methods and show how they can leak information about the selected class. Then, we define distribution-aware methods and show that the maximizer of iAUC is a distribution-aware method. Finally, we introduce two baseline distribution-aware methods that compute Shapley values and are easy to optimize.

\subsection{Class-dependent methods}
\label{sec:class_dep}
Feature attribution methods can be divided into two categories: \emph{class-dependent} and \emph{class-independent}.

We define a class-dependent feature attribution method to be any approach that generates a feature attribution vector as a function of class.
Formally, for each sample of data $x$ and class $c$, a class-dependent feature attribution method $e(x, c): \mathcal{X} \times \mathcal{Y} \to \R^d$ generates an attribution vector such that $e(x, c) \neq e(x,c')$ for some $c \neq c'$.
LIME, SHAP, Grad-CAM, IntGrad, SmoothGrad, and FastSHAP are all examples of class-dependent methods. \Cref{appendix:methods_review} shows how the computation performed by each of these methods is class-dependent.

A class-independent method generates an attribution vector that does not depend on any one class.
Formally, for each sample of data $x$, a class-independent feature attribution method $e(x): \mathcal{X} \to \R^d$ generates an attribution vector as a function of the input $x$.

See \cref{appendix:glossary} for a glossary of terms defined in this paper.

\paragraph{Label leakage.}
In the specific case where a class-dependent method generates an attribution vector using the true label, the predictive performance with only a fixed fraction of features can \emph{exceed} the predictive performance with the entire set of features.
In other words, the class-dependent method is able to leak information about the true label through the feature attributes that is not captured by the full set of features. This leakage would cause the evaluation metric iAUC (\cref{eqn:iauc}) to overestimate the utility of the explanation. Formally,
\begin{lemma}
\label{lemma:leakage}
There exists a class-dependent feature attribution method $e(\rvx,\rvy)$ and data-generating distribution $x,y \sim F(\rvx,\rvy)$ such that
\begin{align}
\label{eq:leakage}
    \E_{F(\rvx, \rvy)}\bigg[\log F(y \g x_{\texttt{top}_n\left(e(x, y)\right)})\bigg] \nonumber \\
    > \E_{ F(\rvx, \rvy)}\bigg[\log F\left(y \g x \right)\bigg]
\end{align}
    for some $n \in [0,100]\%$.
\end{lemma}
The proof can be found in \cref{appendix:proof_leakage}.
\cref{lemma:leakage} shows that the explanation can predict the label better than the full feature set, indicating that the explanations are leaking the label. While Lemma \ref{lemma:leakage} introduces label leakage as a theoretical possibility for class-dependent methods using the true label, we show empirically in \cref{sec:results} that this phenomenon occurs with 
popular class-dependent methods on clinical datasets, up to estimation error of a model trained to approximate $F\left(y \g x \right)$.

\Cref{lemma:leakage} works by having the feature attribution provide low scores to features that reduce the probability of the observed label. Thus, when only considering the top $n\%$ of features, features that reduce the probability of the observed label are obfuscated.
By obfuscating features that support other classes, feature attributions generated by class-dependent methods fail to track the uncertainty of the true label, making the label appear more likely than it should.
This susceptibility could have important implications when interpreting the explanations generated using the true label.
For example, a patient's likelihood of hospital readmission given their discharge summary may only be 55\%, but by omitting the word ``denies'' from a note that reads, ``... pt denies chest pain'' in the discharge summary, the patient may appear to have an 80\% chance of readmission.

\paragraph{Overconfidence using the predicted class.}
As shown in \cref{lemma:leakage}, a feature attribution method should not have access to the true labels when generating feature attributions in order to avoid label leakage. An alternative to using the true label is using a model's prediction of the label.

Let $\hat{y} = \argmax_{y}p_{\text{model}}(y \g x; \theta)$ and let $e'(x, \hat{y})$ be a class-dependent method that uses the model's predicted class. Because $\hat{y}$ is a function of $x$, $e'(x, \hat{y}) = e(x)$, a class-independent method. Therefore, we see that a class-dependent method that uses the predicted class becomes a class-independent method. We call class-dependent methods that use the predicted label \emph{predicted-label-dependent} methods.
Class-independent methods do not leak the label on average:
\begin{lemma}
    \label{lemma:class-independent-no-leakage}
    There does not exist any class-independent feature attribution method $e(\rvx)$ where \cref{eq:leakage} holds for any $F(\rvx,\rvy)$. 
\end{lemma}
The proof can be found in \cref{appendix:proof_noleakage_ind}. 

Predicted-label-dependent methods need not consider the full distribution across all classes. 
They could, for example, focus only on the probability of the predicted class.
The implication is that explanations generated using the predicted class may instead leak the predicted class and omit predictive features that do not support the predicted class. In other words, an explanation could make the predicted class appear more likely than it is  for some subset of feature values. Formally,

\begin{lemma}
\label{lemma:predicted_label}
There exists a predicted-label-dependent feature attribution method $e(\rvx, \hat{y})$ where, for some $x$ where $F(\rvx = x) > 0$ and for some $n \in [0,100]\%$,
\begin{align*}
    F(\rvy=\hat{y} \g x_{\texttt{top}_n\left(e(x,\hat{y})\right)};\beta) &> F(\rvy=\hat{y} \g x).
\end{align*}
\end{lemma}

The proof can be found in \cref{appendix:proof_predicted_label}.

\cref{lemma:predicted_label} demonstrates that an end user runs the risk of drawing false conclusions when interpreting an explanation generated for the predicted class with a class-dependent method.
As an example, consider a model that predicts a patient's likelihood of all-cause mortality to be $52\%$ from data for that patient including clinical notes.
Let us say that a clinician is starting a shift in the hospital, and while they do not have time to read all of the patient's clinical notes, they would like to read the most critical portions of the clinical notes as relates to the patient's likelihood of all-cause mortality.
Now suppose the critical portions of the text are highlighted using a predicted-label-dependent method. Then for some instances, the clinician will miss those features that have a negative relationship with all-cause mortality, but that would still help to inform how they might choose to care for the patient during their shift.

\subsection{Distribution-aware methods}
\label{sec:dist_aware}
The challenge with the full-space of class-independent methods is that class-independent methods need not respect the whole distribution of the label given the inputs, $F(y \g x)$. To limit to methods that consider the whole distribution, we define \emph{distribution-aware} feature attribution methods.

A distribution-aware feature attribution method is a class-independent method $e(x)$ that focuses on the data distribution of the label given the features, $F(y \g x)$. Formally, let $D$ be a probability divergence, and $h(x)$ be a perturbation function. Then for some distribution $r$ a distribution-aware feature attribution method can be written in terms of the divergence $D\Big(F\big(\rvy \g x\big) \mid\mid r\big(\rvy \g h(x)\big)\Big)$ and prefers smaller divergences.
In other words, a distribution-aware method generates feature attributions by measuring the effect of feature perturbation on the distribution of the label. 
The effect is measured by the divergence between the distribution of $\rvy$ given the input and the distribution of $\rvy$ given the perturbed input.
An example perturbation function removes features from the input.
The data distribution $F(y \g x)$ is unavailable, so practical distribution-aware feature attribution methods make use of distributions trained to approximate $F(y \g x)$ such as the surrogate $ p_{\text{surr}}\left(\rvy \g x; \beta\right)$.

How a distribution-aware method prefers a smaller divergence depends on the method. For example, REAL-X \citep{jethani2021explain} is a distribution-aware method that prefers smaller divergences directly through its optimization procedure; we show how the computation performed by REAL-X is distribution-aware in \cref{appendix:methods_review}.

As shown in Lemma \ref{lemma:leakage}, to avoid the potential for label leakage, a feature attribution method should not have access to the true labels when generating feature attributions.
Given the constraint of not using the true labels, we show in \cref{appendix:proof_dist_aware} that the maximizer of iAUC assuming an optimal surrogate $p_{\text{surr}}$ is \emph{not} a class-dependent method, but a distribution-aware method:
\begin{align}
\label{eqn:best_exp_method}
e^* &= \argmin_{e}  \E_{ F(\rvx)} \E_{n \sim \text{Unif}(0,100)}  \nonumber \\ &\bigg[\KL\Big(F\big(\rvy \g x \big) \mid\mid 
    F(\rvy \g x_{\texttt{top}_n\left(e(x)\right)})
    \Big)\bigg].
\end{align}
\cref{eqn:best_exp_method} shows that the optimal feature attribution vector $e^*(x)$ for an instance $x$ is distribution-aware in that it minimizes the KL divergence between the likelihood of the label given all of the features and the likelihood of the target variable given the top $n$\% of the features, averaged across all possible $n$.
Furthermore, we see that $e^*(x)$ does not depend on a true label $y$, but instead averages over a distribution of the label.

The KL divergence, as with many divergences, measures the closeness of two distributions, and thus also measures the calibration in how well the distribution of the target given a subset of features matches the distribution of the target given the full feature set.
Therefore, while a distribution-aware method—like a class-dependent method—returns a subset of the features, the subset that a distribution-aware method returns is calibrated according to the predicted probability. In the all-cause mortality example in \cref{sec:class_dep}, a distribution-aware method would highlight an appropriate ratio of positive and negative features.

\section{DISTRIBUTION-AWARE SHAPLEY VALUE ESTIMATORS}
\label{sec:shap_kl}
Gradient optimization is generally used to solve optimization problems such as the optimal explainer for iAUC.
However, the function $\texttt{top}_n$ in \cref{eqn:best_exp_method} is not differentiable.
We develop two baseline distribution-aware methods, \textbf{SHAP-KL} and \textbf{FastSHAP-KL}, that yield real-valued and, therefore, simpler optimization problems with a squared loss.

SHAP-KL and FastSHAP-KL estimate Shapley values. To compute a Shapley value for each input feature, one must first define how to value a subset of features.
Given \cref{eqn:best_exp_method}, we propose valuing a subset of features according to the KL divergence between the distribution of $\rvy$ given the full set of features and the distribution of $\rvy$ given a subset of the features:
\begin{equation*}
\label{eqn:kl_value}
    v_{x}(s) = -\KL \big( p_{\text{surr}}\left(\rvy \g x; \beta\right) \mid\mid p_{\text{surr}}(\rvy \g m(x, s); \beta) \big).
\end{equation*}
Notice that a higher value for a subset of features entails a smaller KL divergence, as required for distribution-aware methods.

Letting $n \sim \mathcal{U}(\mathcal{D})$ denote a uniform distribution over the set $\mathcal{D}$ of the number of features $\{0,\dots,d-1\}$ to include in a subset, and letting $s \sim \mathcal{U}(\mathcal{P}_i(n))$ denote a uniform distribution over all possible feature subsets (power set of $\{0,1\}^d$) such that $n$ features are included in the subset ($|s|_0 = n$) and the $i$th feature is not included in the subset ($s_i \neq 1$),  the definition of a Shapley value for the $i$th feature is
\begin{align}
    \label{eqn:shap}
    \phi_i(v) &= \E_{n\sim  \mathcal{U}(\mathcal{D})}
    \E_{s \sim \mathcal{U}(\mathcal{P}_i(n))}
    \Big[ v_{x}(s + e_i) - v_{x}(s) \Big] \nonumber \\ 
    &= \E_{n\sim  \mathcal{U}(\mathcal{D})}
    \E_{s \sim \mathcal{U}(\mathcal{P}_i(n))} 
    \E_{y \sim F(\rvy \g x)} \\ 
    &\Big[ \log F\big(y \g m(x, s + e_i)\big) - \log F\big(y \g m(x, s)\big) \Big]. \nonumber
\end{align}
\cref{eqn:shap} shows that this KL divergence-based Shapley value assigns an attribution to a feature based on how much it increases the log probability of the label when added to different subsets of the rest of the features. Note that the maximizer of iAUC (\cref{eqn:best_exp_method}) is a weighted average across subsets that progressively increase in size (e.g. the top $1$\% of features is a strict subset of the top $2$\% of features); the Shapley value (\cref{eqn:shap}) is a weighted average across all possible feature subsets.

Unfortunately, Shapley values introduce computational challenges: the expectation in \cref{eqn:shap} involves an exponential number of subsets, making it infeasible to calculate for large $d$. 
Therefore, SHAP-KL and FastSHAP-KL efficiently approximate the Shapley values.
Following \citet{lundberg2017unified}, SHAP-KL computes Shapley values using its least-squares characterization:
\begin{multline}
\label{eqn:shap_kl}
    e_{\text{SHAP-KL}}(x) = \\
    \argmin_{\phi} \E_{p(\rvs)}\left[\left(v_{x}(s) - s^T\phi - v_{x}(\mathbf{0}) \right)^2 \right].
\end{multline}
Following \citet{jethani2022fastshap}, FastSHAP-KL learns an explanation model $\phi_{\text{fast-kl}}(x; \eta)$ that outputs Shapley values by minimizing the following objective:
\begin{multline}
    \label{eqn:fastshap_kl}
    \mathcal{L}_{\text{FastSHAP-KL}}(\eta) =\\ \E_{F(\rvx)} \E_{p(\rvs)} \Big[ \big( v_{x}(s) - \rvs^\top \phi_{\text{fast-kl}}(x; \eta) - v_{x}(\mathbf{0}) \big)^2 \Big]
\end{multline}
where the feature attributions are generated in a single forward-pass through the explanation model: $e_{\text{FastSHAP-KL}}(x) = \phi_{\text{fast-kl}}(x; \eta)$.
For both objectives (\cref{eqn:shap_kl,eqn:fastshap_kl}), the efficiency constraint and subset sampling distribution $p(\rvs)$ are the same as for SHAP and are presented in \cref{appendix:methods_review}.

\section{EXPERIMENTS}
\label{sec:experiments}
We validate our theoretical findings by performing a comprehensive evaluation of ten of the most commonly used feature attribution methods using three clinical datasets of different high-dimensional data types: biosignals, images, and text. We also compare SHAP-KL to its class-dependent counterpart SHAP-S using the general image dataset CIFAR10 \citep{krizhevsky2009learning}, demonstrating similar findings as in the clinical datasets (\cref{appendix:cifar10}).

\subsection{Datasets and model tasks}
For biosignal data, we use the PTB-XL ECG dataset \citep{wagner2020ptb}. We detect right bundle branch block (RBBB) from ECG inputs using a ResNet model adapted from \citet{hannun2019cardiologist} (we include details of the model architecture in \cref{appendix:ecg_model_arch}). For image data, we use the EyePACs retinal fundus imaging dataset \citep{graham2015kaggle}. We detect the presence and severity of diabetic retinopathy in retinal images using a DenseNet121 model \citep{Huang_2017_CVPR} pre-trained on ImageNet.
For text data, we use the MIMIC-IV critical care dataset \citep{mimiciv}.
We predict 30-day readmission from patients' hospital discharge summaries using the pre-trained Bio+Discharge Summary BERT model \citep{alsentzer2019publicly, huang2019clinicalbert}.
We provide details on dataset processing and splits in \cref{appendix:datasets} and details on training the prediction models in \cref{appendix:prediction_model}.

\subsection{Feature attribution methods}
We evaluate the following seven class-dependent methods: LIME, SHAP, Grad-CAM, IntGrad, SmoothGrad, FastSHAP, and SHAP-S \citep{covert2021explaining,frye2020shapley}.
We evaluate the following three distribution-aware methods: SHAP-KL, REAL-X, and FastSHAP-KL.
Because Grad-CAM was designed for CNNs, we did not evaluate Grad-CAM using MIMIC-IV.
REAL-X failed to optimize on MIMIC-IV using five different regularization hyperparameters, therefore we did not evaluate REAL-X on MIMIC-IV. REAL-X likely requires additional tuning for this task given that it uses score-function gradient optimization.
We provide details on explanation generation in \cref{appendix:explanation_generation}; describe how iAUC is empirically calculated in \cref{appendix:empirical_iauc}; and report training and explanation run-times for each method in \cref{appendix:exp_times}.

\begin{figure*}[t]
    \centering
    \includegraphics[width=.8\linewidth]{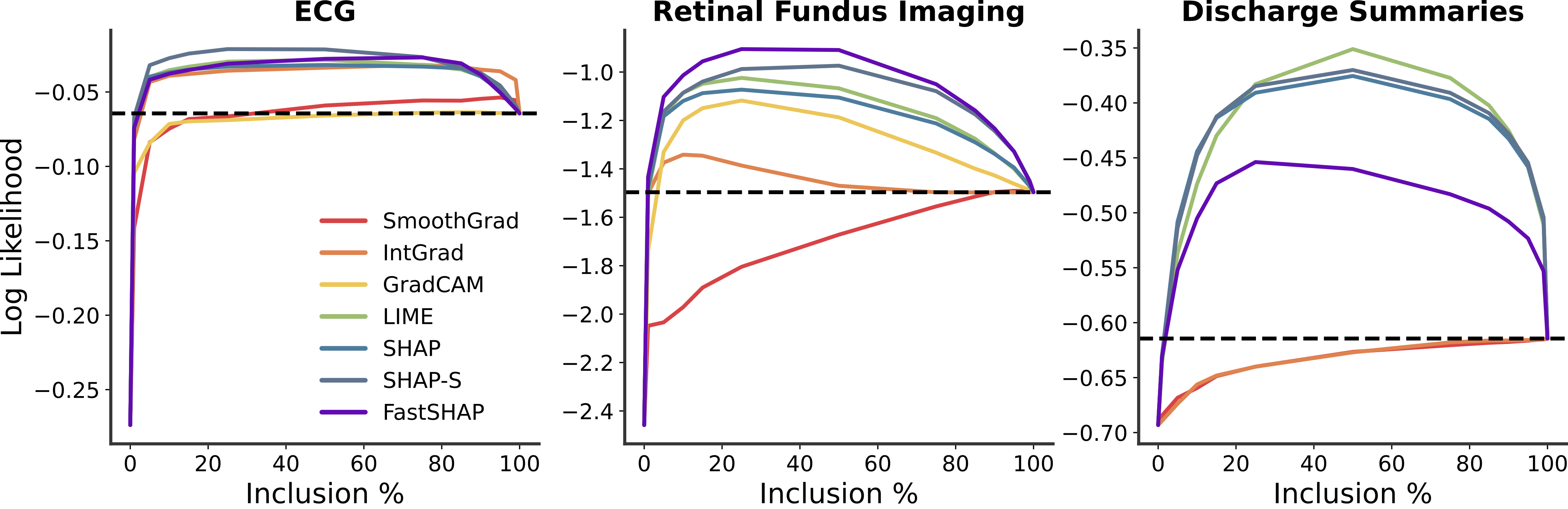}
    \caption{When generating explanations using the true label, class-dependent methods can leak information about the true label that is not captured by the full feature set: performance when using a subset of the most relevant features exceeds performance when using the full feature set (represented by the horizontal dotted lines above).}
    \label{fig:class_dep_curve}
\end{figure*}
\begin{figure*}[t]
    \centering
    \includegraphics[width=.8\linewidth]{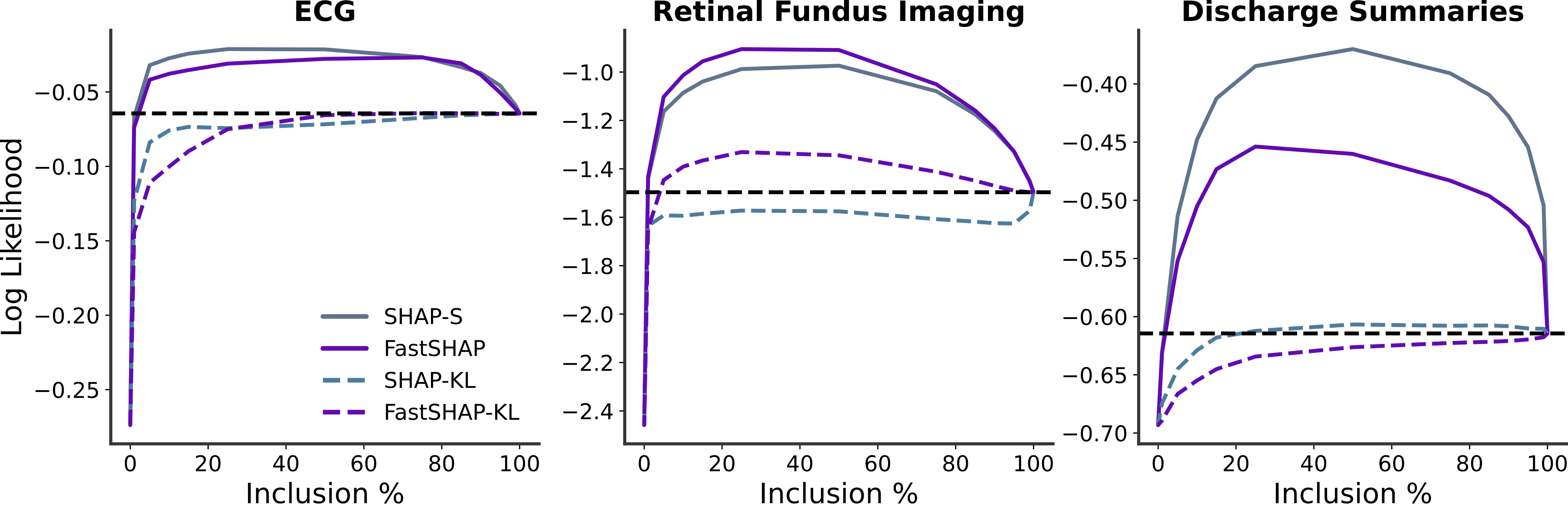}
    \caption{The performance of SHAP-KL and FastSHAP-KL when using a subset of the most relevant features generally does not exceed performance when using the full feature set (represented by the horizontal dotted lines), validating that distribution-aware methods do not leak the label on average.}
    \label{fig:dist_aware_curve}
\end{figure*}
\subsection{Results}
\label{sec:results}
\paragraph{Label leakage in class-dependent methods using the true label.}
First, we plot the log-likelihood inclusion curves of the seven evaluated class-dependent methods when generating an attribution vector using the true label (\cref{fig:class_dep_curve}).
In general, as important features are included in the input to the surrogate evaluation model, the likelihood of the true label (and therefore the log-likelihood across the entire dataset) should increase.
On all three datasets we find that the performance of many of the class-dependent methods when using a subset of the most relevant features exceeds performance when using the full set of features (represented by the horizontal dotted line in \cref{fig:class_dep_curve}).

With finite data and an imperfect surrogate evaluation model $p_{\text{surr}}$, the excess performance could be due to either estimation error or label leakage. 
Therefore, unless we know a priori how the features are related to the input, it is difficult to know whether the unexpectedly high performance of the class-dependent methods is due to label leakage or due to better estimation of the surrogate with fewer features.

\begin{table*}[t]
\caption{Evaluation of the feature attribution methods using iAUC when class-dependent methods use the predicted class. Parentheses indicate 95\% confidence intervals.}
\vspace{-6pt}
\label{tab:auc}
\begin{center}
\begin{tabular}{lccc}
\toprule
                   & \multicolumn{3}{c}{iAUC}                                                                               \\ \cmidrule{2-4} 
                   & PTB-XL: Biosignals               & EyePACs: Images                  & MIMIC-IV: Text                   \\
\midrule
\textit{Distribution-aware} & \multicolumn{1}{l}{}             & \multicolumn{1}{l}{}             & \multicolumn{1}{l}{}             \\
~ ~ FastSHAP-KL        & -0.075 (-0.081, -0.071)          & \textbf{-1.400 (-1.419, -1.386)} & -0.634 (-0.638, -0.631)          \\
~ ~ REAL-X             & \textbf{-0.068 (-0.075, -0.064)} & -1.879 (-1.897, -1.855)          &                                  \\
~ ~ SHAP-KL            & -0.073 (-0.080, -0.066)          & -1.596 (-1.619, -1.578)          & -0.618 (-0.623, -0.613)          \\
\midrule
\textit{Predicted-label-dependent}    & \multicolumn{1}{l}{}             & \multicolumn{1}{l}{}             & \multicolumn{1}{l}{}             \\
~ ~ FastSHAP           & -0.088 (-0.097, -0.082)          & -1.851 (-1.879, -1.825)          & -0.627 (-0.632, -0.623)          \\
~ ~ Grad-CAM           & -0.069 (-0.076, -0.064)          & -1.988 (-2.018, -1.962)          &                                  \\
~ ~ IntGrad            & -0.128 (-0.141, -0.117)          & -1.443 (-1.461, -1.422)          & -0.635 (-0.638, -0.632)          \\
~ ~ LIME               & -0.095 (-0.103, -0.087)          & -1.594 (-1.609, -1.571)          & \textbf{-0.614 (-0.620, -0.609)} \\
~ ~ SHAP               & -0.097 (-0.106, -0.089)          & -1.598 (-1.612, -1.565)          & -0.615 (-0.621, -0.608)          \\
~ ~ SHAP-S             & -0.095 (-0.105, -0.085)          & -1.623 (-1.650, -1.597)          & \textbf{-0.614 (-0.618, -0.607)} \\
~ ~ SmoothGrad         & -0.130 (-0.143, -0.120)          & -1.718 (-1.742, -1.695)          & -0.634 (-0.637, -0.631)         \\
\bottomrule
\end{tabular}
\end{center}
\vspace{-12pt}
\end{table*}
\paragraph{Distribution-aware methods do not demonstrate label leakage.}
Next, we compare our baseline distribution-aware methods SHAP-KL and FastSHAP-KL to their class-dependent counterparts SHAP-S and FastSHAP. We plot the log-likelihood inclusion curves of the four methods using the true label to select which class to explain for SHAP-S and FastSHAP (\cref{fig:dist_aware_curve}).
We find that the performance of SHAP-KL and FastSHAP-KL when using a subset of the most relevant features generally does not exceed performance when using the full set of features, validating that distribution-aware feature attribution methods do not leak the label on average (\cref{lemma:class-independent-no-leakage}).

FastSHAP-KL on the retinal fundus imaging dataset and SHAP-KL on the discharge summaries dataset generate feature attributions that achieve slightly higher log-likelihoods when using a subset of the features than when using the full set of features (\cref{fig:dist_aware_curve}).
Since the performance of a distribution-aware method provably cannot exceed the performance using all features (\Cref{lemma:class-independent-no-leakage}), the amount SHAP-KL and FastSHAP-KL rise above the performance estimate using the full features (the horizontal dotted line) provides a window into the magnitude of relative model misestimation for different subset sizes. This magnitude of model misestimation is smaller than the excess performance over the full feature set in class-dependent methods, suggesting that label leakage, not model estimation, is the primary driver of excess performance in class-dependent methods.

During training, the surrogate evaluation model takes as input a vector of masked features to approximate the probability distribution of the target given a possible subset of features.
It is possible that the surrogate evaluation model is better able to optimize over subsets with fewer features.
Furthermore, since there is an exponential number of subsets, learning to model each conditional distribution given each subset is a difficult task.
However, as a sanity check, we ensure that the surrogate evaluation model performs as well as the original prediction model when evaluated on the full feature set (\cref{appendix:original_vs_eval}).

\paragraph{Predicted-label-dependent vs. distribution-aware methods.}
Finally, we evaluate the iAUC of the ten feature attribution methods when using the predicted class (instead of the true label) to select which class to explain for the seven class-dependent methods (\cref{tab:auc}).

As the most relevant features are included as input to the surrogate evaluation model, we expect the iAUC of a successful feature attribution method to increase.
Though the theory shows that the best method for iAUC is distribution-aware (\cref{eqn:best_exp_method}), the distribution-aware methods studied do not directly optimize iAUC, leaving open the possibility for a predicted-label-dependent method to have higher iAUC. We find that compared to predicted-label-dependent methods, distribution-aware methods have, on average, higher iAUCs on two of the three datasets: REAL-X obtained the highest iAUC (-0.068) on the ECG dataset and FastSHAP-KL obtained the highest iAUC (-1.400) on the retinal fundus imaging dataset. On the discharge summaries dataset, however, the predicted-label-dependent methods outperform the distribution-aware methods on average: LIME and SHAP-S obtained the highest iAUCs (both -0.614).

\section{DISCUSSION}
\subsection{Choosing a feature attribution method}
\label{sec:choosing_method}
When using the true label, distribution-aware methods are recommended given that they do not demonstrate label leakage.
When using the predicted label, however, it is not clear whether a predicted-label-dependent method or a distribution-aware method would be preferred. While in theory a class-dependent method does not perform optimally with respect to iAUC (\cref{eqn:best_exp_method}), it can still outperform a distribution-aware method in practice because existing distribution-aware methods do not optimize iAUC directly (\cref{sec:results}).

In order to evaluate a feature attribution method given some fixed model, we recommend constructing an inclusion curve for the method under consideration as described in \cref{sec:evaluation}. The inclusion curve can then be used to determine how much of the model's performance is explained by different subsets of the top features. For example, an inclusion curve might reveal that the top $10\%$ of features explains $90\%$ of the model's accuracy under some attribution method. If the performance is high enough given the desired percentage of features, the feature attribution method can be used. If it is not high enough, alternative feature attribution methods should be evaluated.

\subsection{The merits of class-dependent methods}
\label{sec:merits_class_dep}
While our theoretical and empirical results demonstrate that class-dependent methods can make a given class appear overly likely, there are settings in which focusing on a single class, instead of on the full distribution across all classes, is a useful design feature (as opposed to a ``bug'') of class-dependent methods.
Because iAUC measures how well the target distribution can be approximated using a subset of features, our paper focuses specifically on settings in which each data point can take on different values of the target distribution (because the true label or predicted class for one sample may not be the same for another sample).

While class-dependent methods do not maximize iAUC and may leak the label, they are still useful when trying to understand which features increase or decrease the probability of a specific class, in which case explanations are generated using a fixed class for \emph{all} data points.
For example, given a model that predicts which molecules inhibit growth of a bacterial species, a class-dependent method might help highlight moieties that maximize the likelihood of that outcome in order to guide molecule development. It remains open what is the best evaluation for a class-specific explanation.

\subsection{The limitations imposed by discrete optimization}
As discussed in \cref{sec:shap_kl}, directly maximizing evaluation metrics for feature attribution methods can be infeasible since they often involve discrete functions that are not differentiable, such as $\texttt{top}_n$ in \cref{eqn:best_exp_method}.
While SHAP-KL and FastSHAP-KL serve as distribution-aware baselines that yield real-valued optimization problems with a squared loss, neither optimizes iAUC directly, which could negatively affect their performances.
The development of additional distribution-aware methods that make use of advances in discrete optimization to more directly optimize evaluation metrics such as \cref{eqn:best_exp_method} is an important avenue for future work.

\subsection{Interpreting the feature attribution vector}
\label{sec:interpret_best_vector}
As discussed in \cref{sec:evaluation}, feature attribution scores can produce many possible explanations, and often it is not known in advance which $n$\% of features will ultimately be of interest. When this is the case, feature attribution method performance can be evaluated across different feature subset sizes and measured using a summary statistic such as iAUC. Eventually, a single feature attribution vector is produced that includes a score for each input feature. Because iAUC is a weighted average, if we were to use the single feature attribution vector to select the ``top'' $k$ features, it is not guaranteed that we would in fact retrieve the most predictive feature subset of size $k$.

To see why, consider a scenario in which there are three input features $x_1$, $x_2$, and $x_3$: \emph{together} $x_1$ and $x_2$ are perfectly predictive of the output, but \emph{separately} they are not very predictive of the output; $x_3$ alone is almost, but not quite, perfectly predictive of the output. Given $k = 1$, the most predictive feature would be $x_3$. Given $k=2$, the most predictive two features would be $x_1$ and $x_2$. However, given the constraint that all features are ranked and the relevant feature subsets monotonically increase in size so that each subset always includes the ``top'' $n$\% of features, there is no way to choose $x_3$ when $k=1$ \emph{and} choose $x_1$ and $x_2$ when $k=2$.

Therefore, there is no single attribution vector with scores for all features such that the $k$ highest ranked features are the most predictive $k$ features for all values of $k$. 
Care should be taken when referring to the features with the top scores in the attribution vector as the ``most predictive'' features.
Future work might investigate ways to address this limitation when developing new attribution methods.

\subsection{Cognitive burden of class-dependent methods}
Given a data point, a class-dependent method produces a set of feature attributions for every possible class. A distribution-aware method, on the other hand, produces for a data point a single set of feature attributions, taking into consideration the full distribution of class probabilities. However, this extra degree of freedom afforded by class-dependent methods comes at a cost.

As discussed in \cref{sec:class_dep}, class-dependent methods can surface features that make the selected class appear more reasonable and obfuscate features that support other classes.
Because class-dependent methods are miscalibrated and fail to adequately capture the uncertainty of a class label, it is important that any end user interpreting the results of a class-dependent method take into consideration not only the explanation generated for the selected class, but also the explanations generated for all other classes.
In other words, the end user runs the risk of drawing inaccurate conclusions by only looking at the explanation for the selected class.
However, considering the feature attributions generated for every class, and then reducing them to a single explanation for the task at hand, constitutes a significant—and perhaps unrealistic—cognitive burden on the part of the end user.
Future work should explore the effect of miscalibrated explanations on human decision-making.

\section{CONCLUSION}
In this work, we introduce and define class-dependent and distribution-aware feature attribution methods.
We demonstrate that class-dependent methods—but not distribution-aware methods—can leak information about the true label, causing evaluation metrics to overestimate the utility of their explanations.
We show that explanations generated by class-dependent methods using the predicted label can make the predicted class appear more likely than it is.
We establish that the maximizer of iAUC is a distribution-aware method.
We present two baseline distribution-aware methods, SHAP-KL and FastSHAP-KL, that can be easily optimized.
Finally, we validate our theoretical findings by evaluating seven class-dependent and three distribution-aware feature attribution methods on three clinical datasets.

\section{REPRODUCIBILITY}
Formal statements and proofs for all theoretical results are provided in \cref{appendix:methods_review,appendix:proof_leakage,appendix:proof_noleakage_ind,appendix:proof_predicted_label,appendix:proof_dist_aware}.
Experimental details for all empirical results are provided in \cref{appendix:cifar10,appendix:ecg_model_arch,appendix:datasets,appendix:prediction_model,appendix:explanation_generation,appendix:empirical_iauc,appendix:exp_times,appendix:original_vs_eval,appendix:exp_architectures} and code is available at \url{https://github.com/explanationleakage/xai}.
All datasets used are publicly available as outlined in \cref{appendix:datasets}.

\subsubsection*{Acknowledgments}
We thank the reviewers for their thoughtful comments.
This work was supported by NIH T32GM007308, NIH T32GM136573, a DeepMind Scholarship, NIH/NHLBI Award R01HL148248, NSF Award 1922658 NRT-HDR: FUTURE Foundations, Translation, and Responsibility for Data Science, and NSF CAREER Award 2145542.

\bibliography{main}

\appendix
\onecolumn
\newpage
\section{FEATURE ATTRIBUTION METHODS GROUPED BY TYPE}
\label{appendix:method_groups}
\begin{table}[h]
\label{tab:method_groups}
\begin{center}
\begin{tabular}{lcccl}
\toprule
                   & Gradient-based       & Removal-based        & Amortized            & Relies on OOD inputs  \\
\midrule
\textit{Distribution-aware} & \multicolumn{1}{l}{} & \multicolumn{1}{l}{} & \multicolumn{1}{l}{} &                       \\
~ ~ FastSHAP-KL        &                      & x                    & x                    &                       \\
~ ~ REAL-X             & \textbf{}            & x                    & x                    &                       \\
~ ~ SHAP-KL            &                      & x                    &                      &                       \\
\midrule
\textit{Class-dependent}    & \multicolumn{1}{l}{} & \multicolumn{1}{l}{} & \multicolumn{1}{l}{} &                       \\
~ ~ FastSHAP           &                      & x                    & x                    &                       \\
~ ~ Grad-CAM           & x                    &                      &                      &                       \\
~ ~ IntGrad            & x                    &                      &                      & \multicolumn{1}{c}{x} \\
~ ~ LIME               &                      & x                    & \textbf{}            & \multicolumn{1}{c}{x} \\
~ ~ SHAP               &                      & x                    &                      & \multicolumn{1}{c}{x} \\
~ ~ SHAP-S             &                      & x                    & \textbf{}            &                       \\
~ ~ SmoothGrad         & x                    &                      &                      & \multicolumn{1}{c}{x} \\
\bottomrule
\end{tabular}
\end{center}
\end{table}

\newpage
\section{REVIEW OF FEATURE ATTRIBUTION METHODS}
\label{appendix:methods_review}

\subsection{Gradient-based feature attribution methods}
Gradient-based methods determine feature importance using gradients with respect to either the input or intermediate representations of the input \citep{ancona2019gradient}.
Gradients explain how sensitive the model's output is to changes in the input or intermediate representation.
Popular gradient-based methods include Grad-CAM \citep{selvaraju2016grad}, Integrated Gradients (IntGrad) \citep{sundararajan2017axiomatic}, SmoothGrad \citep{smilkov2017smoothgrad}, and others \citep{bach2015pixel,shrikumar2017learning}.

SmoothGrad attributes importance based on how sensitive the output is to small changes in the corresponding feature, where the output is smoothed through the introduction of Gaussian noise as follows:
\begin{align*}
    e_{\text{SG}}(x,y) = \frac{1}{n}\sum_{i=1}^n \frac{\partial p_{\text{model}}\left(y \mid x + \gN(0, \sigma^2) ; \theta\right)}{\partial x}.
\end{align*}
IntGrad attributes importance by computing the average gradient to measure the salience of features in the input relative to a reference input $\bar{x}$ as follows:
\begin{align*}
    e_{\text{IG}}(x,y) = (x - \bar{x}) \odot \frac{1}{n}\sum_{i=1}^n \frac{\partial p_{\text{model}}\left(y \mid \bar{x} + \frac{i}{n}(x-\bar{x}) ; \theta\right)}{\partial \left(\bar{x} + \frac{i}{n}(x-\bar{x})\right) }.
\end{align*}
Grad-CAM computes the gradient with respect to an intermediate representation of the input learned by the model $A(x; \theta)$. 
This method can only be used with convolutional neural networks (CNN), as the structure of CNNs uniquely allows the representation to be directly mapped onto the input.
Grad-CAM computes explanations as follows:
\begin{align*}
    e_{\text{Grad-CAM}}(x,y) = \text{ReLu}\left[\sum_{k}^c \left(\frac{1}{hw}\sum_{i-1}^h\sum_{j=1}^w\frac{\partial p_{\text{model}}\left(y \mid x; \theta\right)}{\partial A(x;\theta)_{i,j}^k}\right) A(x;\theta)^k\right]
\end{align*}
where $A(x; \theta)$ is a $c$-channel, $h$ by $w$, two-dimensional convolutional layer.

\subsection{Removal-based feature attribution methods}
Popular removal-based methods include LIME \citep{ribeiro2016should}, SHAP \citep{lundberg2017unified}, and others \citep{zeiler2014visualizing,fong2017interpretable}.

Both LIME and SHAP solve independent optimization problems for each sample of data. SHAP computes Shapley values as follows:
\begin{gather*}
    e_{\text{SHAP}}(x,y) = \argmin_{\phi} \E_{p(\rvs)}\left[ \left(p_{\text{model}}\left(y \mid m(x,s); \theta\right) - s^T\phi
    - F(y) \right)^2 \right]\\
    p(s) \propto \frac{d - 1}{\binom{d}{\mathbf{1}^\top s} \cdot \mathbf{1}^\top s \cdot (d - \mathbf{1}^\top s)} \tag{Shapley kernel}.
\end{gather*}
Similarly, LIME computes feature attributions using a measure of distance $\mathcal{D}$ and attribution complexity $\Omega$ as follows: 
\begin{gather*}
    e_{\text{LIME}}(x,y) = \argmin_{\phi} \E_{p(\rvs)}\left[ \left(p_{\text{model}}\left(y \mid m(x,s); \theta\right) - s^T\phi 
    - F(y) \right)^2 \right] + \Omega(\phi)\\
    p(s) \propto \mathcal{D}(m(x,s), x). \tag{LIME kernel}
\end{gather*}
Both optimizations are performed for a single sample of data $x,y$.
The computation also requires sampling subsets from $p(s)$ and removing input features using a masking function $m(x,s)$ that replaces the removed features with a reference value for high-dimensional data.

Recent work has shown that replacing the removed features with a reference value shifts the input out-of-distribution/off-manifold, which can affect explanation quality and allow for adversarial attack \citep{frye2020shapley,slack2020fooling,jethani2022fastshap}.
To address this issue, SHAP-S \citep{covert2021explaining,frye2020shapley} approximates replacing the removed features $\rvx_{\mathbf{1} - s}$ by using a surrogate model learned by minimizing \cref{eqn:surrogate_objective}.
SHAP-S computes Shapley values as follows:
\begin{gather*}
    e_{\text{SHAP-S}}(x,y) = \argmin_{\phi} \E_{p(\rvs)}\left[ \left(p_{\text{surr}}\left(y \mid m(x,s); \beta\right) - s^T\phi 
    - F(y) \right)^2 \right]\\
    p(s) \propto \frac{d - 1}{\binom{d}{\mathbf{1}^\top s} \cdot \mathbf{1}^\top s \cdot (d - \mathbf{1}^\top s)}.
\end{gather*}
Note that this is the same objective as that used by SHAP except $p_{\text{model}}$ is replaced with $p_{\text{surr}}$.

\subsection{Amortized removal-based feature attribution methods}
The removal-based methods above perform the removal operation for each sample of data, which can be computationally intensive.
\textit{Amortized} removal-based methods represent a new form of removal-based explainability that performs the removal operation across multiple samples of data at a time in order to learn models that produce explanations for a sample of data with a single forward pass \citep{fong2017interpretable,schwab2019cxplain}.
Amortized removal-based methods include L2X \citep{chen2018learning}, INVASE \citep{yoon2018invase}, REAL-X \citep{jethani2021explain}, and FastSHAP \citep{jethani2022fastshap}.

FastSHAP builds upon the SHAP-S objective to learn an explanation model that outputs Shapley values with a single forward pass.
The follow objective is used to train the explanation model:
\begin{gather*}
    \mathcal{L}_{\text{FastSHAP}}(\eta) = \E_{F(\rvx)}\E_{\text{Unif}(\rvy)}\E_{p(\rvs)}\left[ \left(p_{\text{surr}}\left(y \mid m(x,s); \beta\right) - s^T\phi_{\text{explanation}}(x,y; \eta) 
    - F(y) \right)^2 \right]\\
    p(s) \propto \frac{d - 1}{\binom{d}{\mathbf{1}^\top s} \cdot \mathbf{1}^\top s \cdot (d - \mathbf{1}^\top s)}.
\end{gather*}
Explanations can then be computed for a given sample of data as follows:
\begin{align*}
    e_{\text{FastSHAP}}(x,y) = \phi_{\text{explanation}}(x,y; \eta).
\end{align*}

While the above feature attribution methods produce explanations in a class-dependent fashion (as a function of $x$ and $y$), REAL-X produces explanations in a class-independent fashion (as a function of only $x$).
By measuring the KL divergence between the distribution of the target given the full feature set and the distribution of the target given a subset of features, REAL-X is a distribution-aware method.
For a given sample of data, REAL-X returns a sufficiently small subset of features for a given input that best predicts the target and reduces uncertainty about the target variable.
This objective is minimized and amortized by learning an explanation model that outputs a distribution over subsets by sharing parameters across samples of data as follows:
\begin{align*}
    \mathcal{L}_{\text{REAL-X}}(\phi) &= \E_{F(\rvx)}\E_{q_{\text{exp}}(\rvs \mid x; \phi)}\left[\KL \big( F\left(\rvy \mid x; \theta\right) \mid\mid p_{\text{surr}}(\rvy \mid m(x, s); \beta) \right]\\
    &= \E_{F(\rvx)}\E_{q_{\text{exp}}(\rvs \mid x; \phi)}\E_{F\left(y \mid x; \theta\right)}\left[-\log p_{\text{surr}}\left(\rvy \mid m(x, s); \beta\right) \right] + \text{Const.}
\end{align*}
Explanations can then be computed for a given sample of data as follows:
\begin{align*}
    e_{\text{REAL-X}}(x) = q_{\text{exp}}(\rvs \mid x; \phi).
\end{align*}
From this equation, it is clear that REAL-X explanations are generated as a function of $x$ alone.
L2X \citep{chen2018learning} and INVASE \citep{yoon2018invase} optimize a near-identical objective: they learn $p_{\text{surr}}$ jointly within the same objective, which \citet{jethani2021explain} shows allows the predicted distribution to be encoded directly by $s$.

\newpage
\section{GLOSSARY}
\label{appendix:glossary}
\begin{table}[ht]
\begin{tabular}
{p{0.2\linewidth} p{0.75\linewidth}}
\textbf{class-dependent}   & A class-dependent feature attribution method generates a feature attribution vector as a function of class. Formally, for each sample of data $x$ and class $c$, a class-dependent feature attribution method $e(x, c): \mathcal{X} \times \mathcal{Y} \to \R^d$ generates an attribution vector such that $e(x, c) \neq e(x,c')$ for some $c \neq c'$. LIME, SHAP, Grad-CAM, IntGrad, SmoothGrad, and FastSHAP are all examples of class-dependent methods. \Cref{appendix:methods_review} shows how the computation performed by each of these methods is class-dependent. \\\\
\textbf{class-independent} & A class-independent method generates an attribution vector that does not depend on any one class. Formally, for each sample of data $x$, a class-independent feature attribution method $e(x): \mathcal{X} \to \R^d$ generates an attribution vector as a function of the input $x$. \\\\
\textbf{distribution-aware} & A distribution-aware feature attribution method is a class-independent method $e(x)$ that focuses on the data distribution of the label given the features, $F(y \g x)$. Formally, let $D$ be a probability divergence, and $h(x)$ be a perturbation function. Then for some distribution $r$ a distribution-aware feature attribution method can be written in terms of the divergence $D\Big(F\big(\rvy \g x\big) \mid\mid r\big(\rvy \g h(x)\big)\Big)$ and prefers smaller divergences.
In other words, a distribution-aware method generates feature attributions by measuring the effect of feature perturbation on the distribution of the label. 
The effect is measured by the divergence between the distribution of $\rvy$ given the input and the distribution of $\rvy$ given the perturbed input.
An example perturbation function removes features from the input.
The data distribution $F(y \g x)$ is unavailable, so practical distribution-aware feature attribution methods make use of distributions trained to approximate $F(y \g x)$ such as the surrogate $ p_{\text{surr}}\left(\rvy \g x; \beta\right)$. \\\\
&How a distribution-aware method prefers a smaller divergence depends on the method. For example, REAL-X \citep{jethani2021explain} is a distribution-aware method that prefers smaller divergences directly through its optimization procedure; we show how the computation performed by REAL-X is distribution-aware in \cref{appendix:methods_review}. \\\\
\textbf{predicted-label-dependent} & A predicted-label-dependent method is a class-dependent method that uses the predicted label. Let $\hat{y} = \argmax_{y}p_{\text{model}}(y \g x; \theta)$ and let $e'(x, \hat{y})$ be a class-dependent method that uses the model's predicted class. Because $\hat{y}$ is a function of $x$, $e'(x, \hat{y}) = e(x)$, a class-independent method. Therefore, we see that a class-dependent method that uses the predicted class becomes a class-independent method.
\end{tabular}
\end{table}

\newpage
\section{PROOF: LABEL LEAKAGE IN CLASS-DEPENDENT METHODS}
\label{appendix:proof_leakage}
We prove that in the specific case where a class-dependent method generates an attribution vector using the true label, the predictive performance with only a fixed fraction of features can \emph{exceed} the predictive performance with the entire set of features.
In other words, the class-dependent method is able to leak information about the true label through the feature attributes that is not captured by the full set of features. This leakage would cause the evaluation metric iAUC (\cref{eqn:iauc}) to overestimate the utility of the explanation. Formally,
\paragraph{\cref{lemma:leakage}.}
\textit{
    There exists a class-dependent feature attribution method $e(\rvx,\rvy)$ and data-generating distribution $x,y \sim F(\rvx,\rvy)$ such that
\begin{align}
\E_{F(\rvx, \rvy)}\bigg[\log F(y \g x_{\texttt{top}_n\left(e(x, y)\right)})\bigg]
    >
    \E_{ F(\rvx, \rvy)}\bigg[\log F\left(y \g x \right)\bigg] \tag{\ref{eq:leakage}}
\end{align}
    for some $n \in [0,100]\%$.
}

\paragraph{Proof.}
We provide an example scenario in which a class-dependent method identifies a subset of features where the log-likelihood of the target variable using that subset of features exceeds the log-likelihood of the target variable given the full feature set.

Consider the following data-generating process for the input $\rvx:= \{\rvx_1, \rvx_2\}$ and the target $\rvy$:
\begin{align*}
    x_1 \sim \text{Uniform}(0,1), \quad x_2 = \frac{1}{2}, \quad y \sim \text{Bernoulli}\left(\frac{\rvx_1 + \rvx_2}{2}\right),
\end{align*}
and the following class-dependent explanation method:
\begin{gather*}
    e(x, y) = 
    \begin{cases}
        \begin{bmatrix} 0 & 1 \end{bmatrix} & x_1 < 0.5, y=1 \\
        \begin{bmatrix} 1 & 0 \end{bmatrix} & x_1 \geq 0.5, y=1 \\ 
        \begin{bmatrix} 1 & 0 \end{bmatrix} & x_1 \leq 0.5, y=0 \\
        \begin{bmatrix} 0 & 1 \end{bmatrix} & x_1 > 0.5, y=0 \\    
    \end{cases}\\
\end{gather*}
Then for $n=50\%$ (inclusion of a single feature), 
\begin{gather*}
    F(y \g x_{\texttt{top}_{50\%}\left(e(x, y)\right)}) = 
    \begin{cases}
        \frac{1}{2} & x_1 < 0.5, y=1 \\
        \frac{x_1}{2} + \frac{1}{4} & x_1 \geq 0.5, y=1 \\ 
        \frac{3}{4} - \frac{x_1}{2} & x_1 \leq 0.5, y=0 \\
        \frac{1}{2} & x_1 > 0.5, y=0 \\    
    \end{cases}\\
\end{gather*}
Whereas, for the full feature set, 
$$
F\left(y \g x\right) = \begin{cases}
    \frac{x_1}{2} + \frac{1}{4}  &  y=1 \\ 
    \frac{3}{4} - \frac{x_1}{2} &  y=0
\end{cases}
$$
Therefore,
\begin{gather*}
    \begin{matrix}
        F(y \g x_{\texttt{top}_{50\%}\left(e(x, y)\right)}) > F\left(y \g x\right) & \text{if}& x_1 < 0.5, y=1 \\
        F(y \g x_{\texttt{top}_{50\%}\left(e(x, y)\right)}) = F\left(y \g x\right) & \text{if}& x_1 \geq 0.5, y=1 \\ 
        F(y \g x_{\texttt{top}_{50\%}\left(e(x, y)\right)}) = F\left(y \g x\right) &\text{if}&  x_1 \leq 0.5, y=0 \\
        F(y \g x_{\texttt{top}_{50\%}\left(e(x, y)\right)}) > F\left(y \g x\right) & \text{if}& x_1 > 0.5, y=0 \\    
    \end{matrix}.\\
\end{gather*}
Since in all cases $F(y \g x_{\texttt{top}_{50\%}\left(e(x, y)\right)}) \geq F\left(y \g x\right)$ and the events $x_1 < 0.5, y=1$ and $x_1 > 0.5, y=0$ occur with non-zero probability,
$$
\E_{ F(\rvx, \rvy)}\bigg[\log F(y \g x_{\texttt{top}_{50\%}\left(e(x, y)\right)})\bigg]
> \E_{ F(\rvx, \rvy)}\big[\log F\left(y \g x \right)\big].
$$

\newpage
\section{PROOF: NO LABEL LEAKAGE IN CLASS-INDEPENDENT METHODS}
\label{appendix:proof_noleakage_ind}
We showed in \cref{appendix:proof_leakage} that there exists a class-dependent feature attribution method $e(\rvx,\rvy)$ and data-generating distribution $x,y \sim F(\rvx,\rvy)$ such that
\begin{align}
\E_{F(\rvx, \rvy)}\bigg[\log F(y \g x_{\texttt{top}_n\left(e(x, y)\right)})\bigg]
    >
    \E_{ F(\rvx, \rvy)}\bigg[\log F\left(y \g x \right)\bigg] \tag{\ref{eq:leakage}}
\end{align}
    for some $n \in [0,100]\%$.

We now prove that there does not exist any class-independent method $e(\rvx)$ where the likelihood of the target variable given the top $n$\% of features exceeds the likelihood of the target variable given the full feature set.

\paragraph{\cref{lemma:class-independent-no-leakage}.}
\textit{
There does not exist any class-independent feature attribution method $e(\rvx)$ where \cref{eq:leakage} holds for any $F(\rvx,\rvy)$.
}
\paragraph{Proof.}
For a class-independent method \cref{eq:leakage} becomes
\begin{align}
\E_{F(\rvx, \rvy)}\bigg[\log F(y \g x_{\texttt{top}_n\left(e(x)\right)})\bigg]
    >
    \E_{ F(\rvx, \rvy)}\bigg[\log F\left(y \g x \right)\bigg] \label{eq:class-indep-leakage}
\end{align}

The generative process by which the explanations are created using a class-independent method can be expressed via the following Markov chain:
$$
\rvy \xrightarrow{} \rvx \xrightarrow{} \rvx_{\texttt{top}_n\left(e(\rvx)\right)}, 
$$
where the target generates the input, which in turn generates the explanation that is used to mask/include features. According to the data processing inequality,
\begin{gather*}
    I(\rvx; \rvy) \geq 
    I( \rvx_{\texttt{top}_n\left(e(\rvx)\right)}; \rvy) \quad \forall n\in [0,100]\%,
\end{gather*}
which states that the mutual information content between the input and the target cannot be increased by processing the input.
Rewriting the mutual information in terms of the conditional entropy produces
$$
 H(\rvy \mid \rvx) \leq  H(\rvy \mid \rvx_{\texttt{top}_n\left(e(\rvx)\right)})
 \quad \forall n\in [0,100]\%.
$$

Let $r'$ be a distribution. Then using the definition of conditional entropy the inequality can be written as an expectation:
{\small
\begin{align*}
    \E_{ F(\rvx, \rvy)}\big[\log F\left(y \mid x \right)\big] &\geq \E_{F\big(\rvx_{\texttt{top}_n\left(e(\rvx)\right)}, \rvy\big)}\bigg[\log F\Big(y \mid x_{\texttt{top}_n\left(e(x)\right)}\Big)\bigg]\\
    &\geq \E_{F\big(\rvx_{\texttt{top}_n\left(e(\rvx)\right)}, \rvy\big)}\bigg[\log F\Big(y \mid x_{\texttt{top}_n\left(e(x)\right)}\Big)\bigg] \\
        & \quad - \E_{F\big(\rvx_{\texttt{top}_n\left(e(\rvx)\right)}\big)}\bigg[\KL \bigg( F\Big(y \mid x_{\texttt{top}_n\left(e(x)\right)}\Big) \mid\mid r'\Big(y \mid x_{\texttt{top}_n\left(e(x)\right)}\Big) \bigg)\bigg]\\
    &= \E_{F\big(\rvx_{\texttt{top}_n\left(e(\rvx)\right)}, \rvy\big)}\bigg[\log F\Big(y \mid x_{\texttt{top}_n\left(e(x)\right)}\Big)\bigg] \\
        & \quad + \E_{F\big(\rvx_{\texttt{top}_n\left(e(\rvx)\right)}, \rvy\big)}\bigg[\log r'\Big(y \mid x_{\texttt{top}_n\left(e(x)\right)}\Big) - \log F\Big(y \mid x_{\texttt{top}_n\left(e(x)\right)}\Big) \bigg]\\
    &= \E_{F\big(\rvx_{\texttt{top}_n\left(e(\rvx)\right)},\rvy\big)}
    \bigg[\log r'\Big(y \mid x_{\texttt{top}_n\left(e(x)\right)}\Big)\bigg]\\
    &= \E_{F\big(\rvx_{\texttt{top}_n\left(e(\rvx)\right)},\rvy\big)}
    \E_{F\big(\rvx \mid \rvx_{\texttt{top}_n\left(e(\rvx)\right)}, \rvy\big)} \bigg[
    \log r'\Big(y \mid x_{\texttt{top}_n\left(e(x)\right)}\Big) \bigg]\\
    &= \E_{F\big(\rvx_{\texttt{top}_n\left(e(\rvx)\right)},\rvx,\rvy\big)}
    \bigg[
    \log r'\Big(y \mid x_{\texttt{top}_n\left(e(x)\right)}\Big) \bigg]\\
    &= \E_{F\big(\rvx,\rvy\big)}\E_{F\big(\rvx_{\texttt{top}_n\left(e(\rvx)\right)}|\rvx,\rvy\big)}
    \bigg[
    \log r'\Big(y \mid x_{\texttt{top}_n\left(e(x)\right)}\Big) \bigg]\\
    &= \E_{F\big(\rvx,\rvy\big)}
    \bigg[
    \log r'\Big(y \mid x_{\texttt{top}_n\left(e(x)\right)}\Big) 
    \bigg] & \forall n\in [0,100]\%.
\end{align*}}Since $r'$ is arbitrary, \cref{eq:class-indep-leakage} holds for $r'=F$. Therefore, the performance of a class-independent method when using the top $n$\% of features is upper-bounded by the performance when using the full feature set.

\newpage
\section{PROOF: LABEL LEAKAGE IN CLASS-DEPENDENT METHODS WHEN USING THE PREDICTED CLASS}
\label{appendix:proof_predicted_label}
Predicted-label-dependent methods need not consider the full distribution across all classes. 
They could, for example, focus only on the probability of the predicted class.
The implication is that explanations generated using the predicted class may instead leak the predicted class and omit predictive features that do not support the predicted class. In other words, an explanation could make the predicted class appear more likely than it is  for some subset of feature values. Formally,
\paragraph{\cref{lemma:predicted_label}.}
There exists a predicted-label-dependent feature attribution method
$e(\rvx, \hat{y})$ where, for some $x$ where $F(\rvx = x) > 0$ and for some $n \in [0,100]\%$,
\begin{align*}
    F(\rvy=\hat{y} \g x_{\texttt{top}_n\left(e(x,\hat{y})\right)};\beta) &> F(\rvy=\hat{y} \g x).
\end{align*}

\paragraph{Proof.}
Let $\rvx, \rvy \sim F(\rvx, \rvy)$ be the data-generating distribution. Consider the following data-generating process for the input $\rvx:= \{\rvx_1, \rvx_2\}$ and the target $\rvy \in \{1, 2, 3\}$:
\begin{gather*}
    \rvx_1 \sim \text{Bernoulli}(0.80), \quad \rvx_2 \sim \text{Bernoulli}(0.5),\\
    \rvy \g \rvx \sim \text{Categorical}([\max\{\frac{\rvx_1-\rvx_2}{2}, 0\}+\frac{1}{2}, \frac{1-(\max\{\frac{\rvx_1-\rvx_2}{2}, 0\}+\frac{1}{2})}{2}, \frac{1-(\max\{\frac{\rvx_1-\rvx_2}{2}, 0\}+\frac{1}{2})}{2}])
        .
\end{gather*}
Let the model's predicted class be $\hat{y} = \argmax\limits_{\rvy}p_{\text{model}}\left(y \g x; \theta\right)$ where $p_{\text{model}}$ is the optimal model (i.e. the true data-generating distribution). Notice that it will always be the case that $\hat{y} = 1$.
Let $e(x,\hat{y}=1)$ be a class-dependent feature attribution method that uses the predicted class to generate explanations and that is defined as follows:
\begin{align*}
    e(\rvx,\hat{y}=1) = \begin{cases}
        \begin{bmatrix} 1 & 0 \end{bmatrix} & \text{if } \rvx_1 = 1\\
        \begin{bmatrix} 0 & 1 \end{bmatrix} & \text{if } \rvx_1 = 0
        \end{cases}.
\end{align*}
Then for $n=50\%$ (inclusion of a single feature), 
\begin{align*}
    F(\rvy=\hat{y} \g \rvx_{\texttt{top}_{50\%}\left(e(\rvx,\hat{y})\right)};\beta) = 
    \begin{cases}
        F(\rvy = \hat{y} \g \rvx_1 = 1) = 0.5 \cdot 0.5 + 0.5 \cdot 1.0 = 0.75 & \text{if } \rvx_1 = 1, \rvx_2 = 1\\
        F(\rvy = \hat{y} \g \rvx_1 = 1) = 0.5 \cdot 0.5 + 0.5 \cdot 1.0 = 0.75 & \text{if } \rvx_1 = 1, \rvx_2 = 0\\
        F(\rvy = \hat{y} \g \rvx_2 = 1) = 0.8 \cdot 0.5 + 0.2 \cdot 0.5 = 0.5 & \text{if } \rvx_1 = 0, \rvx_2 = 1\\
        F(\rvy = \hat{y} \g \rvx_2 = 0) = 0.8 \cdot 1.0 + 0.2 \cdot 0.5 = 0.9 & \text{if } \rvx_1 = 0, \rvx_2 = 0
    \end{cases}\\
\end{align*}
and
\begin{align*}
    F(\rvy=\hat{y} \g \rvx) = 
    \begin{cases}
        0.5 & \text{if } \rvx_1 = 1, \rvx_2 = 1\\
        1.0 & \text{if } \rvx_1 = 1, \rvx_2 = 0\\
        0.5 & \text{if } \rvx_1 = 0, \rvx_2 = 1\\
        0.5 & \text{if } \rvx_1 = 0, \rvx_2 = 0
    \end{cases}.\\
\end{align*}
We see that
\begin{align*}
    F(\rvy=\hat{y} \g \rvx_{\texttt{top}_{50\%}\left(e(\rvx,\hat{y})\right)};\beta) &> F(\rvy=\hat{y} \g \rvx) \quad \text{if } \rvx_1 = 1, \rvx_2 = 1\\
    F(\rvy=\hat{y} \g \rvx_{\texttt{top}_{50\%}\left(e(\rvx,\hat{y})\right)};\beta) &> F(\rvy=\hat{y} \g \rvx) \quad \text{if } \rvx_1 = 0, \rvx_2 = 0.
\end{align*}
Since both of the above two cases occur with non-zero probability, we see that there exists a class-dependent feature attribution method that, when generating explanations using the predicted class, makes the predicted class appear overly likely for some $n \in [0,100]\%$.

\newpage
\section{PROOF: WITHOUT TRUE LABELS, A DISTRIBUTION-AWARE METHOD MAXIMIZES IAUC}
\label{appendix:proof_dist_aware}
Given the constraint of not using the true labels, we show that the maximizer of iAUC assuming an optimal surrogate $p_{\text{surr}}$ is \emph{not} a class-dependent method, but a distribution-aware method.

As we saw in \cref{eqn:iauc}, the area under the inclusion curve (iAUC) is
\begin{gather*}
    \text{iAUC} = \E_{n \sim \text{Unif}(0,100)}\E_{ F(\rvx, \rvy)}\Bigg[\log p_{\text{surr}}\Big(y \g 
    m\big(x, \texttt{top}_n\left(e(x,y)\right); \beta \big)
    \Big)\Bigg].
\end{gather*}
The feature attribution method that depends \textit{only} on $x$ and that maximizes iAUC is
\begin{align*}
    e^* &= \argmax_{e} \E_{n \sim \text{Unif}(0,100)}\E_{ F(\rvx, \rvy)}\Bigg[\log p_{\text{surr}}\Big(y \g 
    m\big(x, \texttt{top}_n\left(e(x)\right); \beta \big)
    \Big)\Bigg]\\
    &=\argmax_{e} \E_{n \sim \text{Unif}(0,100)}\E_{ F(\rvx)} \E_{F(\rvy \g \rvx)}\Bigg[\log p_{\text{surr}}\Big(y \g 
    m\big(x, \texttt{top}_n\left(e(x)\right); \beta \big)
    \Big)\Bigg]\\
    &=\argmax_{e} \E_{n \sim \text{Unif}(0,100)}\E_{ F(\rvx)} \E_{F(\rvy \g \rvx)}\Bigg[\log p_{\text{surr}}\Big(y \g 
    m\big(x, \texttt{top}_n\left(e(x)\right); \beta \big)
    \Big) - \log p_{\text{surr}}\big(y \g x; \beta \big) + \log p_{\text{surr}}\big(y \g x; \beta \big) \Bigg]\\
    &=\argmax_{e} \E_{n \sim \text{Unif}(0,100)}\E_{ F(\rvx)} \E_{F(\rvy \g \rvx)}\Bigg[\log p_{\text{surr}}\Big(y \g 
    m\big(x, \texttt{top}_n\left(e(x)\right); \beta \big)
    \Big) - \log p_{\text{surr}}\big(y \g x; \beta \big) \Bigg]\\
    &=\argmin_{e} \E_{n \sim \text{Unif}(0,100)}\E_{ F(\rvx)} \E_{F(\rvy \g \rvx)}\Bigg[\log p_{\text{surr}}\Big(y \g x; \beta \Big) - \log p_{\text{surr}}\Big(y \g 
    m\big(x, \texttt{top}_n\left(e(x)\right); \beta \big)
    \Big) \Bigg]
\end{align*}
Since at optimality $p_{\text{surr}}\Big(\rvy \g m\big(\rvx, \texttt{top}_n\left(e(x)\right); \beta^* \big)\Big) = F(\rvy \g \rvx_{\texttt{top}_n\left(e(x)\right)})$ for all $n \in [0,100]\%$ \citep{jethani2021explain}, we see that
\begin{align*}
    e^* &= \argmin_{e}  \E_{ F(\rvx)} \E_{n \sim \text{Unif}(0,100)} \Bigg[\KL\bigg(F\big(\rvy \g x \big) \mid\mid 
    F\big(\rvy \g x_{\texttt{top}_n\left(e(x)\right)}\big)
    \bigg)\Bigg].
\end{align*}
Therefore, we see that the optimal feature attribution vector $e^*(x)$ for an instance $x$ is distribution-aware in that it minimizes the KL divergence between the likelihood of the label given all of the features and the likelihood of the target variable given the top $n$\% of the features, averaged across all possible $n$.
Furthermore, we see that $e^*(x)$ does not depend on a true label $y$, but instead averages over a distribution of the label.

\newpage
\section{COMPARING SHAP-KL AND SHAP-S ON CIFAR-10}
\label{appendix:cifar10}
We compare the distribution-aware SHAP-KL to its class-dependent counterpart SHAP-S using the general image dataset CIFAR10 \citep{krizhevsky2009learning}, demonstrating similar findings as in the clinical datasets.

CIFAR-10 contains $60{,}000$ $32\times32$ images across $10$ classes.
We used $50{,}000$ samples for the training and $5{,}000$ samples for both validation and testing.
Each image was resized to $224 \times 224$ using bilinear interpolation to interface with the ResNet-50 architecture \citep{he2015deep}.
The ResNet models were trained as described in \cref{appendix:prediction_model}.
Explanations were then generated for the 5,000 images in the test set using SHAP-KL and SHAP-S (both set up to sample 4028 feature subsets), and evaluated using iAUC.

\begin{figure*}[h]
    \centering
    \includegraphics[width=.4\linewidth]{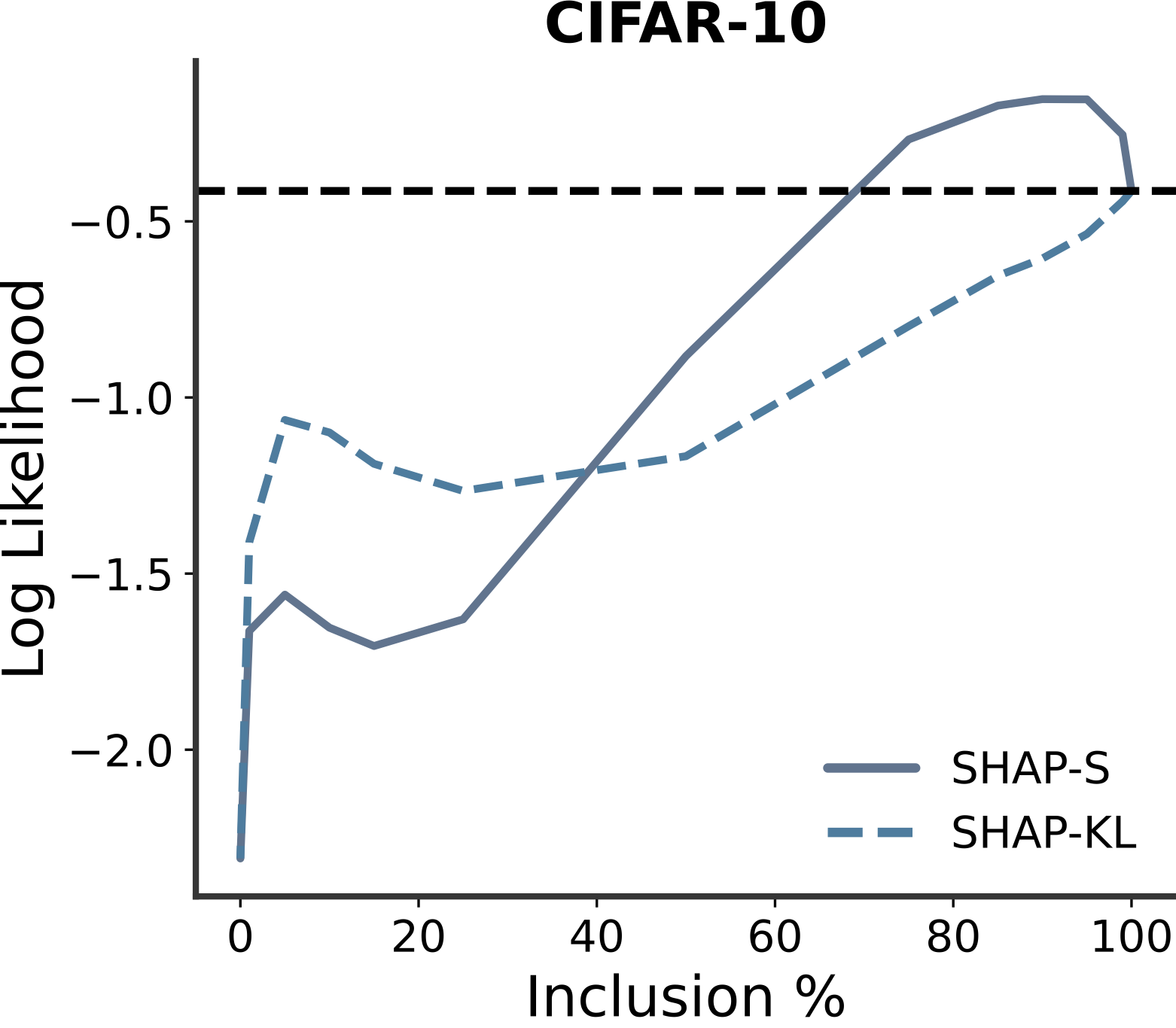}
    \caption{The performance of SHAP-KL when using a subset of the most relevant features does not exceed performance when using the full set of features (represented by the horizontal dotted line), validating that distribution-aware methods do not leak the label on average.}
    \label{fig:cifar_leakage}
\end{figure*}

As with our other experiments (\cref{sec:results}), we plot the log-likelihood inclusion curves of SHAP-KL and SHAP-S, using the true label for SHAP-S (\cref{fig:cifar_leakage}).
We find that the performance of SHAP-KL when using a subset of the features does not exceed performance when using the full set of features (represented by the horizontal dotted line), validating that distribution-aware methods do not leak the label on average.
We find that the performance of SHAP-S when using a subset of the features does exceed performance when using the full set of features.

We also evaluate the iAUC of SHAP-KL and of SHAP-S when using the predicted class (instead of the true label) to select which class to explain (\cref{tab:cifar10}).
We find that SHAP-KL has a higher iAUC than SHAP-S.

\begin{table}[h]
    \centering
    \caption{Evaluation of SHAP-KL and SHAP-S using iAUC when SHAP-S uses the predicted class.}
    \begin{tabular}{lcc}
        \toprule
	& SHAP-KL & SHAP-S \\\midrule
        iAUC & \textbf{-1.008} & -1.461\\
        \bottomrule
    \end{tabular}
    \label{tab:cifar10}
\end{table}

\newpage
\section{ECG PREDICTION MODEL ARCHITECTURE}
\label{appendix:ecg_model_arch}
\begin{figure}[h]
    \centering
    \includegraphics[width=0.7\linewidth]{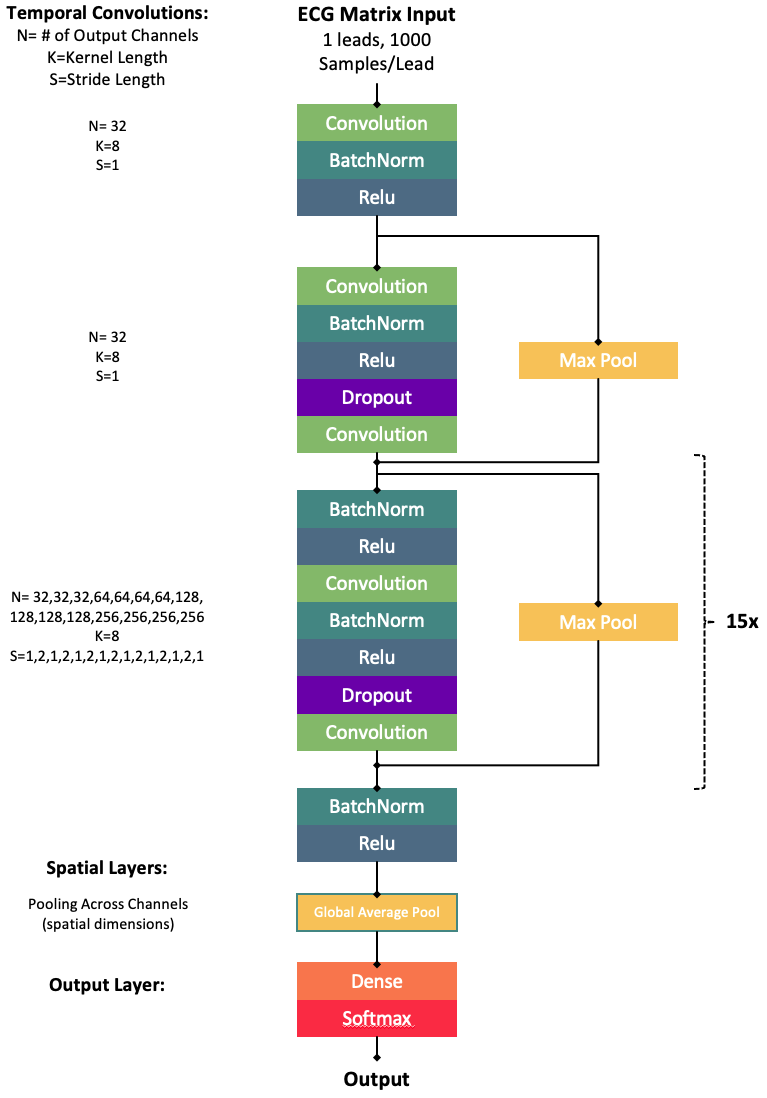}
    \caption{Diagram of the ECG model architecture.}
    \label{fig:ecg_model_arch}
\end{figure}

\newpage
\section{DATASET DETAILS}
\label{appendix:datasets}
The PTB-XL dataset consists of $21,430$ $12$-Lead/$10$s ECGs.
We classify the presence of right bundle branch block (RBBB) from lead VI of the ECG.
There was a $7.75\%$ prevalence of RBBB in the dataset. 
The dataset was split into training, validation, and test sets according to an $8:1:1$ ratio. 
The validation and test sets were directly used as the explanation validation and test sets ($2,163$ ECGs).
We trained the deeper $34$-layer ResNet model adapted from \citet{hannun2019cardiologist} to classify the presence of a RBBB (see \cref{appendix:ecg_model_arch} for model architecture) as compared to shallower architectures \citep{goodfellow2018towards, han2020deep}.
The PTB-XL data is made available under the Creative Commons Attribution 4.0 International Public License at \url{https://physionet.org/content/ptb-xl/1.0.2/}.

The EyePACs dataset consists of $88,702$ retinal fundus images.
The task is formulated as multiclass classification labeled for the presence and severity of diabetic retinopathy.
The label distribution was $73.5\%$ normal, $7\%$ mild, $15\%$ moderate, $2.5\%$ severe, and $2\%$ proliferative. 
The dataset was split into training, validation, and test sets according to a $4:1:5$ ratio. 
The dataset was downloaded from Kaggle and processed with TensorFlow Datasets \citep{tensorflow2015-whitepaper} according to the $2015$ Kaggle competition winner \citet{graham2015kaggle} to generate $544$ by $544$ pixel images. 
Class-balanced explanation validation and test sets (1000 and 2500 images, respectively) were randomly sampled from the validation and test sets, respectively. 
We trained a DenseNet121 model, pre-trained on ImageNet, to classify the severity of diabetic retinopathy.
The EyePACs dataset is made available at \url{https://www.kaggle.com/c/diabetic-retinopathy-detection/data} under a set of rules found here: \url{https://www.kaggle.com/competitions/diabetic-retinopathy-detection/rules}.

We processed the MIMIC-IV dataset according to \citet{huang2019clinicalbert}, yielding a cohort of $34,560$ patient admissions with $2,963$ positive 30-day readmission labels and $42,358$ negative labels.
As detailed by \citet{huang2019clinicalbert}, the dataset was balanced to yield a final dataset of $5,926$ discharge summaries and split into training, validation, and test sets according to an $8:1:1$ ratio.
The validation and test sets were directly used as the explanation validation and test sets ($584$ discharge summaries).
The discharge summaries were split into $128$ token segments and tokenized using the Hugging Face \citep{wolf2019huggingface} BERT tokenizer. 
We trained a BERT transformer, implemented using Hugging Face for TensorFlow and pretrained weights obtained from \citet{alsentzer2019publicly}, to predict $30$-day readmission.
The MIMIC-IV data is made available under the PhysioNet Credentialed Health Data License 1.5.0 at \url{https://physionet.org/content/mimiciv/2.0/}.

\newpage
\section{PREDICTION MODEL DETAILS}
\label{appendix:prediction_model}
Both the DenseNet121 and ResNet models were trained for 100 epochs using Adam with a learning rate of $10^{-3}$ and a batch size of $32$. 
The BERT model was trained for $50$ epochs using Adam with a learning rate of $2*10^{-5}$ and a batch size of $16$.
We used a learning rate scheduler that multiplied the learning rate by a factor of $0.95$ after three epochs of no validation loss improvement. 
Early stopping was triggered after the validation loss ceased to improve for ten epochs.

\newpage
\section{EXPLANATION GENERATION}
\label{appendix:explanation_generation}

The explanation validation set was used to tune each explanation method's hyperparameter, where the best performing explanation set for the associated metric was selected.
The gradient-based methods were implemented using TensorFlow's native backprop functionality. 
Both IntGrad and SmoothGrad have a single hyperparameter that controls the number of samples along the path from the baseline input to the final input and the number of noisy inputs to sample, respectively.
We tuned this hyperparamater for both explanation methods across $64,128,256,512,$ and $1024$ samples. 
Grad-CAM does not have any hyperparameters. Because Grad-CAM can only be applied to CNNs, we could not generate Grad-CAM explanations for our BERT text classification model. 

Feature removal was performed at different granularities for the removal-based methods depending on the data type.
We removed segments of the input instead of individual features: $32$ by $32$ super-pixel segments for images, $0.08$ second segments, and tokens for text.
We implement LIME and SHAP using their respective open-source packages, where feature removal is simulated by replacing the removed features with a baseline value for all three data types.
We simply alter SHAP's value function to implement SHAP-S and SHAP-KL.
Image and ECG data were replaced using the zero baseline, while text data was replaced using the \texttt{[MASK]} token. 
Both LIME, SHAP, SHAP-S, and SHAP-KL have a single hyperparameter that controls the number of feature subsets to sample.
We tuned this hyperparamater for the explanation methods across $512,1024,2048,4096,$ and $8192$ subset samples. 
LIME\footnote{\url{https://github.com/marcotcr/lime}} is made available under the BSD 2-Clause ``Simplified'' License.
SHAP\footnote{\url{https://github.com/slundberg/shap}} is made available under the MIT License.

Explanation generation with REAL-X, FastSHAP, and FastSHAP-KL involves a three-step process: 1) training a surrogate model to simulate feature removal, 2) training an explanation model, and 3) computing explanations with a single forward pass through the explanation model.
We trained the surrogate and explanation models and tuned them using the same training and validation sets we used to train the original model.
The surrogate model was trained with random feature removals, where the removed features were replaced with their aforementioned baseline.
The surrogate model simulates marginalizing out features from the original model with their conditional distribution.
The surrogate model's training procedure and model architecture directly mirrored that of the corresponding original model.
We adapted the explanation model architectures from their corresponding classification models being explained by truncating the architectures (see \cref{appendix:exp_architectures} for details).
For FastSHAP and FastSHAP-KL, we tuned the hyperparameter controlling the number of feature subsets to sample for each input in a mini-batch explanation model across $1,2,4,8,16$.
For REAL-X, we tuned the regularization hyperparameter that enforces sparse subset selections of the input across $10^{-5},10^{-4},10^{-3},10^{-2},$ and $10^{-1}$.
All other training hyperparameters used to train the original models were conserved when training and tuning the explanation models.
FastSHAP\footnote{\url{https://github.com/neiljethani/fastshap}} and REAL-X\footnote{\url{https://github.com/rajesh-lab/realx}} are both made available under the MIT License.

\newpage
\section{EMPIRICAL CALCULATION OF IAUC}\label{appendix:empirical_iauc}

As discussed in \cref{sec:evaluation}, an inclusion curve is constructed by progressively increasing $n$ from 0 to 100, selecting the top $n$\% of features for each data point in a held-out test set using the corresponding feature attribution vector $e(x,y)$, and then measuring performance of the surrogate evaluation model $p_{\text{surr}}\Big(\rvy \g m\big(x, \texttt{top}_n(e(x,y)) \big);\beta\Big)$ across the entire held-out test set using the log-likelihood.
As we saw in \cref{eqn:iauc}, the area under the inclusion curve (iAUC) is
\begin{gather*}
    \text{iAUC} = \E_{n \sim \text{Unif}(0,100)}\E_{ F(\rvx, \rvy)}\Bigg[\log p_{\text{surr}}\Big(y \g 
    m\big(x, \texttt{top}_n\left(e(x,y)\right); \beta \big)
    \Big)\Bigg].
\end{gather*}

For our experiments (\cref{sec:experiments}), instead of calculating the above expectation, we construct the inclusion curve using the following values of $n$: $0,\ 1,\ 5,\ 10,\ 15,\ 25,\ 50,\ 75,\ 85,\ 90,\ 95,\ 99,$ and $100$. 
We then use the trapezoid rule to approximate the area under this inclusion curve in order to calculate iAUC.

\newpage
\section{EXPLANATION EFFICIENCY}\label{appendix:exp_times}
We include wall clock times for each explanation method on the explanation test set below.
The experiments were run using a single core of an Intel Xeon Gold 6148 processor and a single AMD MI50 GPU.
Gradient-based methods are quite efficient, especially Grad-CAM, which only requires a single gradient estimate per explanation. Removal-based methods such as SHAP and LIME are slow. Meanwhile, the amortized methods incur a fixed training cost, which is often made up for by its meager marginal cost for generating each explanation. In some cases, the amortization also improves explanation quality. However, this may not always be the case, as the model may over-fit the data (i.e. FastSHAP on MIMIC-IV) or optimize poorly.
\begin{table*}[h]
\label{tab:runtime}
    \begin{center}
    \caption{Training and explanation run-times for the explanation sets (in minutes).}
    \begin{tabular}{llccccc}
    \toprule
     & & PTB-XL & EyePACs & MIMIC-III\\
    \midrule
    & \textit{\# of Samples} & 2163 & 2500 & 3063 \\
    \midrule
     \multirow{10}{*}{\rotatebox[origin=c]{90}{Explain}}
    & Grad-CAM & 56.39 & 31.11 & ---\\
    & IntGrad & 15.82 & 3902.53 & 717.18\\
    & SmoothGrad & 63.17 & 2398.13 & 415.74\\
    & SHAP & 143.02 & 6389.12 & 1343.03\\
    & SHAP-S & 150.49 & 6909.33 & 1354.13\\
    & SHAP-KL & 144.47 & 6348.32 & 1356.01\\
    & LIME & 138.08 & 6649.19 & 1340.14\\
    & FastSHAP & 0.01 & 0.66 & 0.30\\
    & FastSHAP-KL & 0.02 & 0.51 & 0.27\\
    & REAL-X & 0.01 & 0.51 & ---\\
    \midrule
    \multirow{5}{*}{\rotatebox[origin=c]{90}{Train}}
    & FastSHAP & 54.67 & 3597.14 & 2038.53\\
    & FastSHAP-KL & 76.51 & 2918.45 & 3320.96\\
    & REAL-X & 89.27 & 3611.14 & ---\\
    & SHAP-S & 16.73 & 1320.97 & 38.63\\
    & SHAP-KL & 16.73 & 1320.97 & 38.63\\
    \bottomrule
    \end{tabular}
    \end{center}
\end{table*}

\newpage
\section{SURROGATE MODEL VS. ORIGINAL PREDICTION MODEL PERFORMANCE}
\label{appendix:original_vs_eval}
\begin{table}[h]
    \centering
    \caption{\textbf{AUROC of the original prediction model compared to the surrogate model.} The prediction model is trained using the full feature set while the surrogate model is trained using random subsets of the input. The performance of each model on the full feature set is compared using the AUROC (micro-averaged for Eye-PACs). Model performance is negligibly affected by randomly removing subets of the input during training.}
    \begin{tabular}{lcc}
        \toprule
         & $p_{\text{model}}(\rvy \mid \rvx; \theta)$ & $p_{\text{surr}}(\rvy \mid \rvx; \alpha)$ \\\midrule
        PTB-XL & 0.997 & 0.997\\
        Eye-PACs & 0.947 & 0.951\\
        MIMIC-IV & 0.775 & 0.774\\
        \bottomrule
    \end{tabular}
    \label{tab:orginal_vs_eval}
\end{table}

\newpage
\section{EXPLANATION MODEL ARCHITECTURES}
\label{appendix:exp_architectures}
\subsection{ECG explanation model}
We modified the ECG model architecture (see \cref{appendix:ecg_model_arch}) to return a tensor of size $125 \times 1$ for REAL-X/FastSHAP-KL and  $125 \times 2$ (one for each class) for FastSHAP. 
For the 10s ECG's input size of $1000 \times 1$, this process provides $0.08$ second segment explanations.
First, the layers after the $6^{th}$ residual connection were removed; the output of this block was  $125 \times 64$. 
We then appended a 1D convolutional layer with filters of size 1 × 1, one filter for REAL-X/FastSHAP-KL and 2 filters for FastSHAP, such that the output was $125 \times 1$ or $125 \times 2$ respectively.
For FastSHAP, the $y^{th}$ $125$ dimensional array slice corresponded to the segment-level Shapley values for the  class $y \in \{0,1\}$.

\subsection{Retinal fundus image explanation model}
We modified the DenseNet121 architecture to return a tensor of size $17 \times 17 \times 1$ for REAL-X/FastSHAP-KL and  $17 \times 17 \times 5$ (one for each class) for FastSHAP. 
For an input image size of $544 \times 544$ this process provides $32 \times 32$ super-pixel explanations.
First, the classification layers (global average pooling and fully-connected layers) were removed; the output of this block was  $17 \times 17 \times 1024$. 
We then appended a 2D convolutional layer with filters of size 1 × 1, one filter for REAL-X/FastSHAP-KL and 5 filters for FastSHAP, such that the output is $17 \times 17 \times 1$ or $17 \times 17 \times 5$ respectively.
For FastSHAP, the $y^{th}$ $17 \times 17$ slice corresponded to the superpixel-level Shapley values for the  class $y \in \{0, 1, 2, 3, 4\}$.

\subsection{Discharge summary explanation model}
We modified the BERT architecture by appending two fully connected layers to the output for the last encoder layer.
The output of the final encoder layer was a $128 \times 768$ tensor, such that there was a $768$ dimensional array outputted for each input token. 
We first appended a fully connected layer with $768$ units, GeLU activation, and layer norm.
Then, we appended another fully connected layer with either $1$ unit for REAL-X/FastSHAP-KL or $2$ units for FastSHAP, yielding a $128 \times 1$ or $128 \times 2$ output.
This output provided attributions for each token in the input text segment.
For FastSHAP, the $y^{th}$ $128$ dimensional slice corresponded to the token-level Shapley values for the  class $y \in \{0, 1\}$.

\vfill

\end{document}